\documentclass{article}

\usepackage{microtype}
\usepackage{graphicx}
\usepackage{booktabs} % for professional tables
\usepackage[space]{grffile}

\usepackage{xparse}   % http://ctan.org/pkg/xparse
\usepackage{amsmath}
\usepackage{amsfonts}
\usepackage{amssymb}
\usepackage{amsthm}
\usepackage{makecell}
\usepackage{caption}
\usepackage{subcaption}
\usepackage{threeparttable}
\usepackage[utf8]{inputenc}
\usepackage{natbib}
\usepackage{bibunits}
\usepackage{appendix}
\usepackage{xcolor}
\usepackage{color,soul}
\usepackage{chngcntr}
\usepackage{caption}

\usepackage{hyperref}

\usepackage[accepted]{icml2018}

\newcommand{\beq}{\begin{equation}}
\newcommand{\eeq}{\end{equation}}

\newcommand{\beqa}{\begin{eqnarray}}
\newcommand{\eeqa}{\end{eqnarray}}

\newcommand{\beqan}{\begin{eqnarray*}}
\newcommand{\eeqan}{\end{eqnarray*}}

\newcommand{\E}{\mathbb{E}}

\newlength{\minipagewidth}
\let\R\undefined %sometimes it is defined as something I don't know
\newcommand{\R}{\mathbb{R}}

\newcommand{\eqdef}{\stackrel{\rm def}{=}}

\newtheorem{predefinition}{Definition}

\newtheorem{theorem}{Theorem}

\newtheorem{preproposition}{Proposition}

\newtheorem{remark}{Remark}

\renewcommand{\phi}{\varphi}
\renewcommand{\epsilon}{\varepsilon}
\newcommand{\eps}{\epsilon}

\renewcommand{\R}{{\cal R}}

\usepackage[accepted]{icml2018}

\icmltitlerunning{IMPALA: Importance Weighted Actor-Learner Architectures}

\begin{document}

\NewDocumentCommand{\rot}{O{45} O{1em} m}{\makebox[#2][r]{\rotatebox{#1}{#3}}}%

\twocolumn[
\icmltitle{IMPALA: Scalable Distributed Deep-RL with Importance Weighted Actor-Learner Architectures}

\icmlsetsymbol{equal}{*}    

\begin{icmlauthorlist}
\icmlauthor{Lasse Espeholt}{equal,dm}
\icmlauthor{Hubert Soyer}{equal,dm}
\icmlauthor{Remi Munos}{equal,dm}
\icmlauthor{Karen Simonyan}{dm}
\icmlauthor{Volodymyr Mnih}{dm}
\icmlauthor{Tom Ward}{dm}
\icmlauthor{Yotam Doron}{dm}
\icmlauthor{Vlad Firoiu}{dm}
\icmlauthor{Tim Harley}{dm}
\icmlauthor{Iain Dunning}{dm}
\icmlauthor{Shane Legg}{dm}
\icmlauthor{Koray Kavukcuoglu}{dm}
\end{icmlauthorlist}

\icmlaffiliation{dm}{DeepMind Technologies, London, United Kingdom}

\icmlcorrespondingauthor{Lasse Espeholt}{lespeholt@google.com}

\icmlkeywords{Machine Learning, DeepRL, Deep Learning, Reinforcement Learning, Large Scale Reinforcement Learning, Deep Reinforcement Learning}

\vskip 0.3in
]

\printAffiliationsAndNotice{\icmlEqualContribution} % otherwise use the standard text.
\begin{abstract}

In this work we aim to solve a large collection of tasks using a single reinforcement learning agent with a single set of parameters. A key challenge is to handle the increased amount of data and extended training time. We have developed a new distributed agent IMPALA (Importance Weighted Actor-Learner Architecture) that not only uses resources more efficiently in single-machine training but also scales to thousands of machines without sacrificing data efficiency or resource utilisation. We achieve stable learning at high throughput by combining decoupled acting and learning with a novel off-policy correction method called V-trace.
We demonstrate the effectiveness of IMPALA for multi-task reinforcement learning on DMLab-30 (a set of 30 tasks from the DeepMind Lab environment \cite{beattie2016dmlab}) and Atari-57 (all available Atari games in Arcade Learning Environment \cite{bellemare13arcade}). Our results show that IMPALA is able to achieve better performance than previous agents with less data, and crucially exhibits positive transfer between tasks as a result of its multi-task approach. The source code is publicly available at \href{https://github.com/deepmind/scalable_agent}{github.com/deepmind/scalable\_agent}.

\end{abstract}

\section{Introduction}
\label{sec:introduction}

Deep reinforcement learning methods have recently mastered a wide variety of domains through trial and error learning \cite{mnih15human,silver2017mastering,silver2016alphago, zoph2017learning, lillicrap2015continuous, maron2018distributional}. While the improvements on tasks like the game of Go \cite{silver2017mastering} and Atari games \cite{horgan2018distributed} have been dramatic, the progress has been primarily in single task performance, where an agent is trained on each task separately. We are interested in developing new methods capable of mastering a diverse set of tasks simultaneously as well as environments suitable for evaluating such methods.

One of the main challenges in training a single agent on many tasks at once is scalability. Since the current state-of-the-art methods like A3C \cite{A3C2016} or UNREAL \cite{jaderberg2016reinforcement} can require as much as a billion frames and multiple days to master a single domain, training them on tens of domains at once is too slow to be practical.

\begin{figure}
    \centering
    \includegraphics[scale=.3]{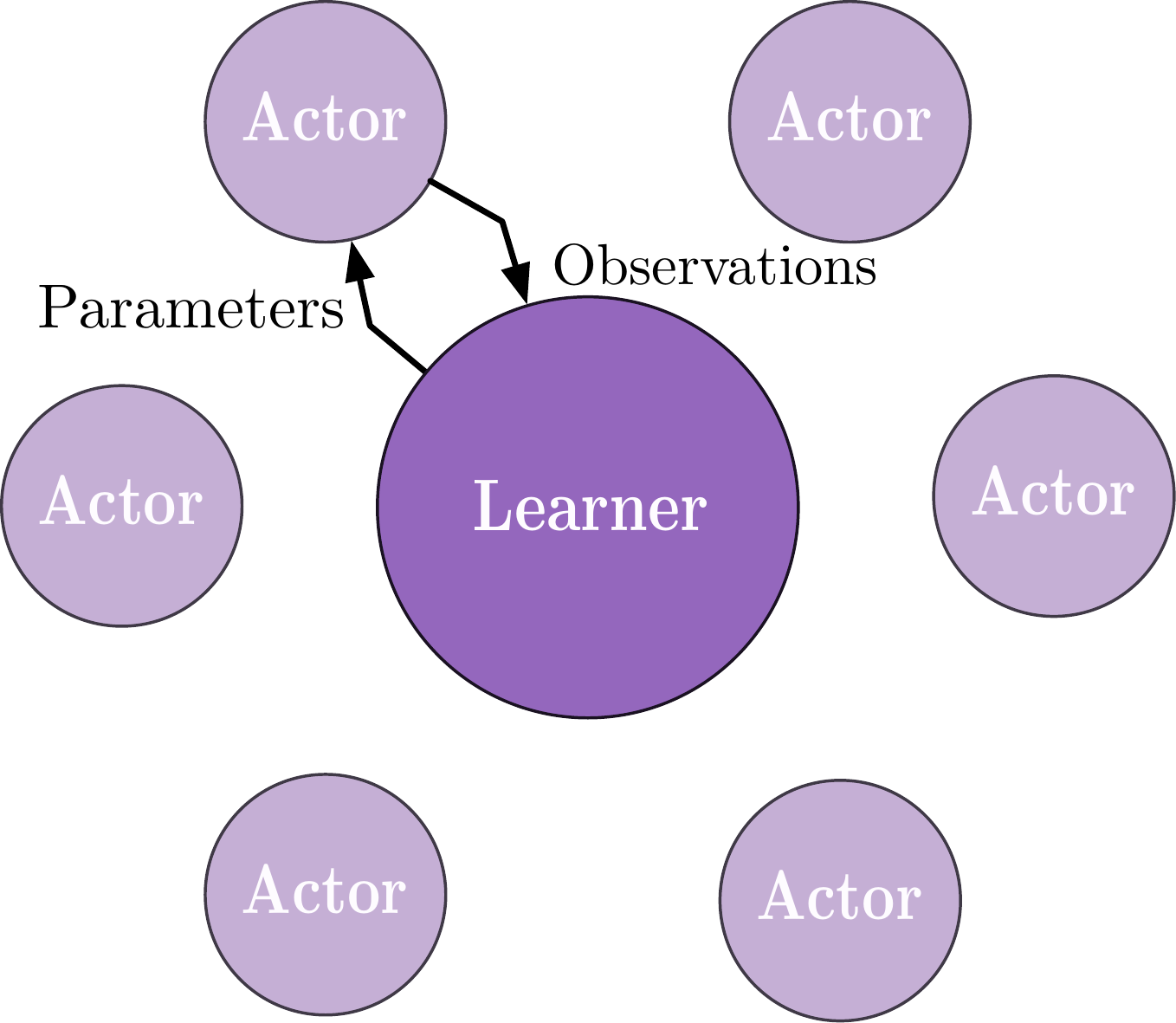}
    \hfill
    \vrule
    \hfill
    \includegraphics[scale=.3]{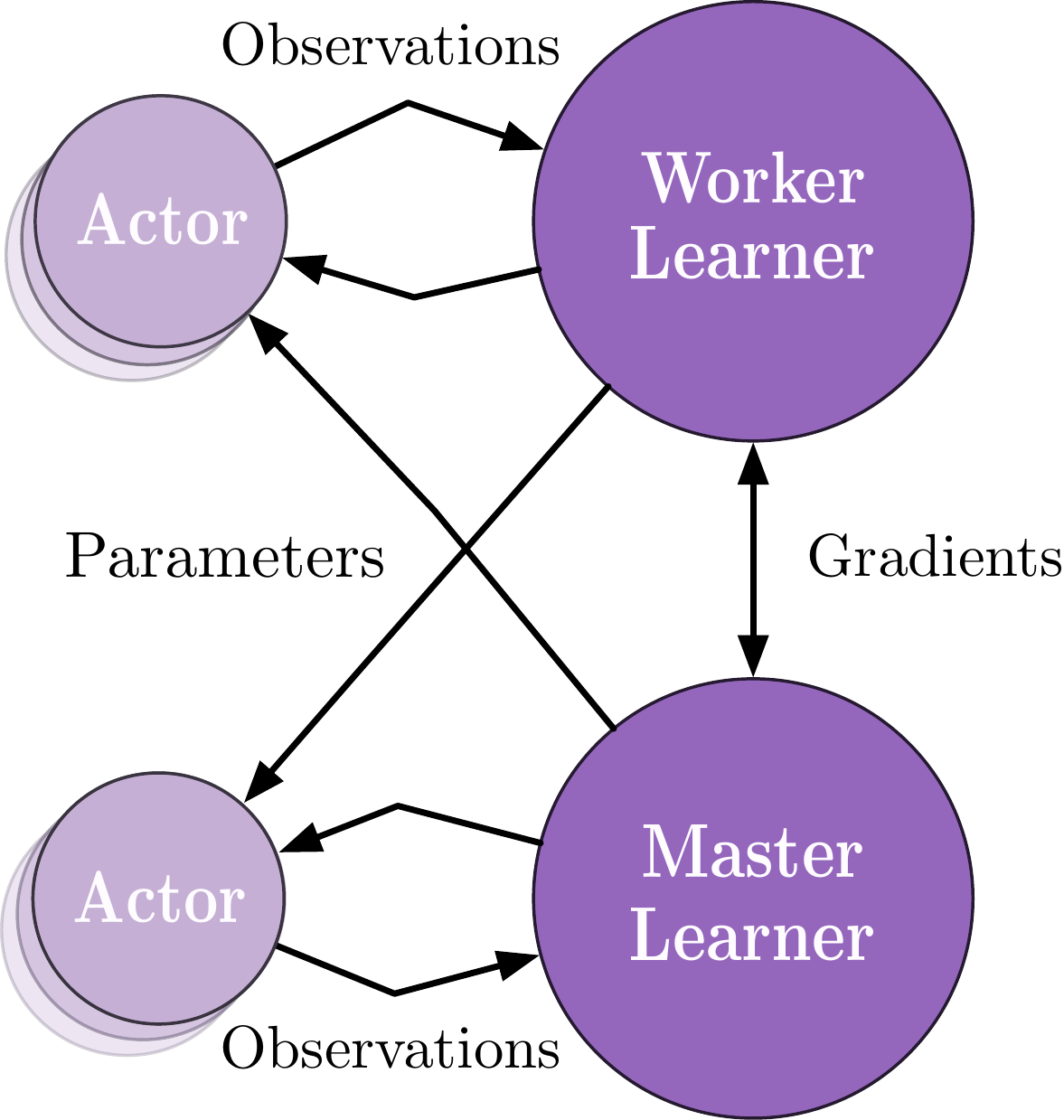}
    \caption {{\bf Left: Single Learner.} Each \emph{actor} generates trajectories and sends them via a queue to the \emph{learner}. Before starting the next trajectory, \emph{actor} retrieves the latest policy parameters from \emph{learner}. {\bf Right: Multiple Synchronous Learners.} Policy parameters are distributed across multiple \emph{learners} that work synchronously.}
    \label{multiple_learners}
    \label{single_learner}
    \label{fig:impala_architecture}
\end{figure}

We propose the \textbf{Imp}ortance Weighted \textbf{A}ctor-\textbf{L}earner \textbf{A}rchitecture (IMPALA) shown in Figure~\ref{fig:impala_architecture}. IMPALA is capable of scaling to thousands of machines without sacrificing training stability or data efficiency.
Unlike the popular A3C-based agents,
in which workers communicate gradients with respect to the parameters of the policy to a central parameter server,
IMPALA actors communicate trajectories of experience (sequences of states, actions, and rewards) to a centralised learner.
Since the learner in IMPALA has access to full trajectories of experience we use a GPU to perform updates on mini-batches of trajectories while aggressively parallelising all time independent operations. This type of decoupled architecture can achieve very high throughput. However, because the policy used to generate a trajectory can lag behind the policy on the learner by several updates at the time of gradient calculation, learning becomes off-policy. Therefore, we introduce the {\em V-trace} off-policy actor-critic algorithm to correct for this harmful discrepancy.

With the scalable architecture and V-trace combined, IMPALA achieves exceptionally high data throughput rates of 250,000 frames per second, making it over 30 times faster than single-machine A3C.
Crucially, IMPALA is also more data efficient than A3C based agents and more robust to hyperparameter values and network architectures, allowing it to make better use of deeper neural networks.
We demonstrate the effectiveness of IMPALA by training a single agent on multi-task problems using DMLab-30, a new challenge set which consists of 30 diverse cognitive tasks in the 3D DeepMind Lab \cite{beattie2016dmlab} environment and by training a single agent on all games in the Atari-57 set of tasks.

\section{Related Work}
\label{sec:related_work}

The earliest attempts to scale up deep reinforcement learning relied on distributed asynchronous SGD \cite{NIPS2012_4687} with multiple workers.
Examples include distributed A3C \cite{A3C2016} and Gorila \cite{DBLP:journals/corr/NairSBAFMPSBPLM15}, a distributed version of Deep Q-Networks \cite{mnih15human}. Recent alternatives to asynchronous SGD for RL include using evolutionary processes \cite{salimans2017evolution}, distributed BA3C \cite{DBLP:journals/corr/abs-1801-02852} and Ape-X \cite{horgan2018distributed} which has a distributed replay but a synchronous learner.

\begin{figure}
    \centering
    \includegraphics[width=.7\columnwidth]{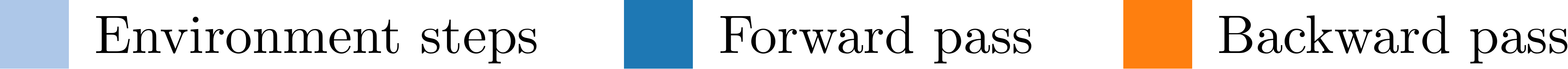}
    \begin{minipage}[c][2.75cm][t]{.22\textwidth}
    \begin{subfigure}{1\textwidth}
        \centering
        \includegraphics[scale=.11]{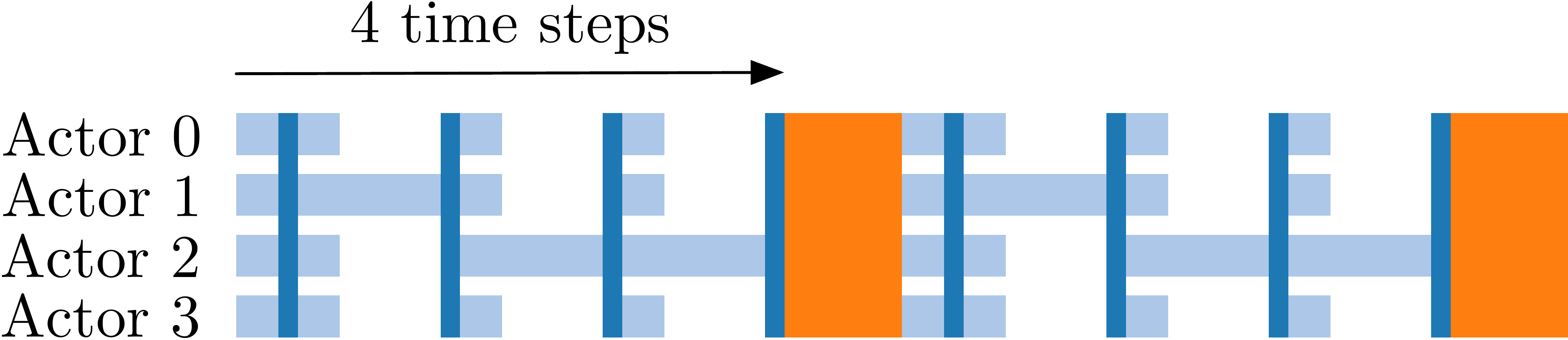}
        \caption{Batched A2C (sync step.)}
        \label{batched_a2c_timeline}
    \end{subfigure}\\
    \begin{subfigure}{1\textwidth}
        \centering
        \includegraphics[scale=.11]{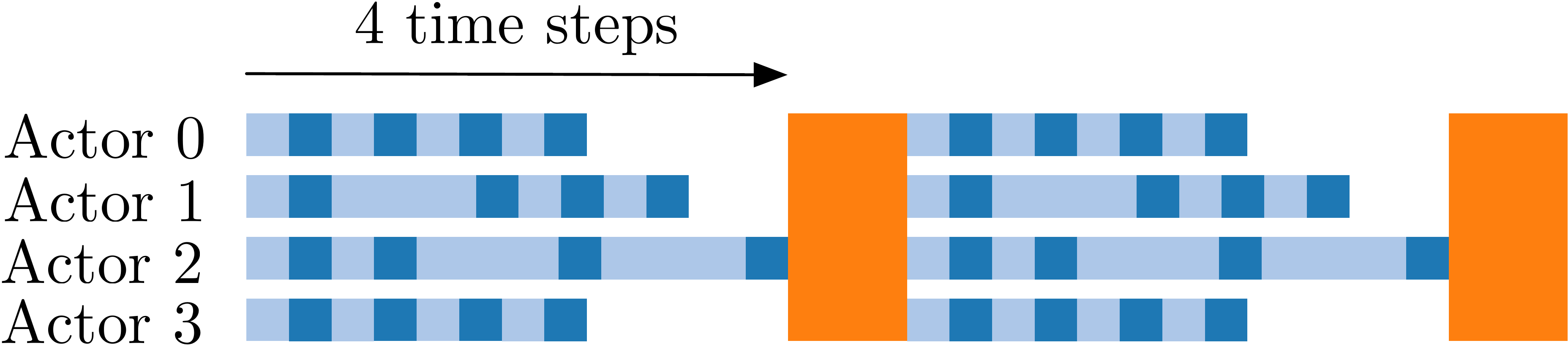}
        \caption{Batched A2C (sync traj.)}
        \label{batched_a2_1c_timeline}
    \end{subfigure}%
    \end{minipage}%
    \begin{minipage}[c][3cm][t]{.22\textwidth}
    \begin{subfigure}{1.2\textwidth}
        \centering
        \includegraphics[scale=.11]{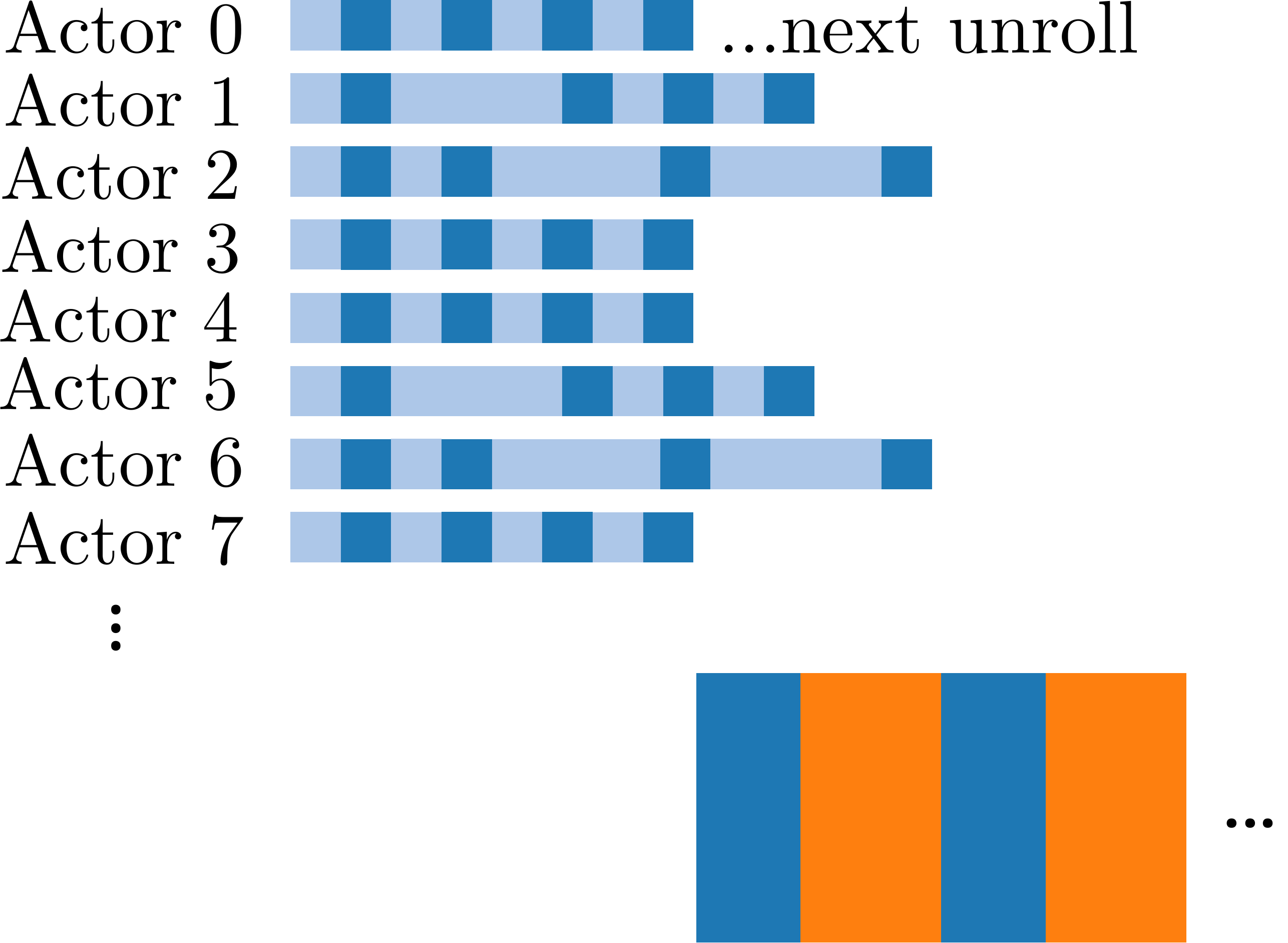}
        \caption{IMPALA}
        \label{IMPALA_ga3c_timeline}
    \end{subfigure}\\
    \end{minipage}
    \caption{Timeline for one unroll with 4 steps using different architectures. Strategies shown in \textbf{(a)} and \textbf{(b)} can lead to low GPU utilisation due to rendering time variance within a batch. In \textbf{(a)}, the actors are synchronised after every step. In \textbf{(b)} after every $n$ steps. IMPALA \textbf{(c)} decouples acting from learning.
    }
\end{figure}

There have also been multiple efforts that scale up reinforcement learning by utilising GPUs. One of the simplest of such methods is batched A2C \cite{DBLP:journals/corr/ClementeMC17}.
At every step, batched A2C produces a batch of actions and applies them to a batch of environments.
Therefore, the slowest environment in each batch determines the time it takes to perform the entire batch step (see Figure \ref{batched_a2c_timeline} and \ref{batched_a2_1c_timeline}). In other words, high variance in environment speed can severely limit performance.
Batched A2C works particularly well on Atari environments, because rendering and game logic are computationally very cheap in comparison to the expensive tensor operations performed by reinforcement learning agents.
However, more visually or physically complex environments can be slower to simulate and can have high variance in the time required for each step. Environments may also have variable length (sub)episodes causing a slowdown when initialising an episode.

The most similar architecture to IMPALA is GA3C \cite{babaeizadeh2016ga3c}, which also uses asynchronous data collection to more effectively utilise GPUs. It decouples the acting/forward pass from the gradient calculation/backward pass by using dynamic batching. The actor/learner asynchrony in GA3C 
leads to instabilities during learning, which \cite{babaeizadeh2016ga3c} only partially mitigates by adding a small constant to action probabilities during the estimation of the policy gradient.
In contrast, IMPALA uses the more principled V-trace algorithm.

Related previous work on off-policy RL include \cite{precup2000eligibility,precup01offpolicy,Wawrzynski09,geist2014off,o2016pgq} and \cite{harutyunyan16qlambda}.
The closest work to ours is the Retrace algorithm  \cite{munos2016safe} which introduced an off-policy correction for multi-step RL, and has been used in several agent architectures \cite{Ziyu2017,Reactor2017}. Retrace requires learning state-action-value functions $Q$ in order to make the off-policy correction.
However, many actor-critic methods such as A3C learn a state-value function $V$ instead of a state-action-value function $Q$.
V-trace is based on the state-value function.
\section{IMPALA}
\label{sec:architecture}

IMPALA (Figure~\ref{fig:impala_architecture}) uses an actor-critic setup to learn a policy $\pi$ and a baseline function $V^{\pi}$.  The process of generating experiences is decoupled from learning the parameters of $\pi$ and $V^{\pi}$. The architecture consists of a set of actors, repeatedly generating trajectories of experience,
and one or more learners that use the experiences sent from actors to learn $\pi$ off-policy.

At the beginning of each trajectory, an actor updates its own local policy $\mu$ to the latest learner policy $\pi$ and runs it for $n$ steps in its environment. After $n$ steps, the actor sends the trajectory of states, actions and rewards  $x_1, a_1, r_1, \dots, x_n, a_n, r_n$ together with the corresponding policy distributions $\mu(a_t|x_t)$ and initial LSTM state to the learner through a queue. The learner then continuously updates its policy $\pi$ on batches of trajectories, each collected from many actors. This simple architecture enables the learner(s) to be accelerated using GPUs and actors to be easily distributed across many machines. However, the learner policy $\pi$ is potentially several updates ahead of the actor's policy $\mu$ at the time of update, therefore there is a \textit{policy-lag} between the actors and learner(s). V-trace corrects for this lag to achieve extremely high data throughput while maintaining data efficiency. Using an actor-learner architecture, provides fault tolerance like distributed A3C but often has lower communication overhead since the actors send observations rather than parameters/gradients.

With the introduction of very deep model architectures, the speed of a single GPU is often the limiting factor during training.
IMPALA can be used with distributed set of learners to train large neural networks efficiently as shown in Figure~\ref{multiple_learners}.
Parameters are distributed across the learners and actors retrieve the parameters from all the learners in parallel while only sending observations to a single learner.
IMPALA use synchronised parameter update which is vital to maintain data efficiency when scaling to many machines \cite{DBLP:journals/corr/ChenMBJ16}.

\subsection{Efficiency Optimisations}
\label{sec:efficiency_optimizations}

GPUs and many-core CPUs benefit greatly from running few large, parallelisable operations instead of many small operations.
Since the learner in IMPALA performs updates on entire batches of trajectories, it is able to parallelise more of its computations than an online agent like A3C.
As an example, a typical deep RL agent features a convolutional network followed by a Long Short-Term Memory (LSTM) \cite{LSTM} and a fully connected output layer after the LSTM. An IMPALA learner applies the convolutional network to all inputs in parallel by folding the time dimension into the batch dimension.
Similarly, it also applies the output layer to all time steps in parallel once all LSTM states are computed.
This optimisation increases the effective batch size to thousands.
LSTM-based agents also obtain significant speedups on the learner by exploiting the network structure dependencies and operation fusion~\cite{DBLP:journals/corr/AppleyardKB16}.

Finally, we also make use of several off the shelf optimisations available in TensorFlow \cite{abadi2017tensorflow} such as preparing the next batch of data for the learner while still performing computation, compiling parts of the computational graph with XLA (a TensorFlow Just-In-Time compiler) and optimising the data format to get the maximum performance from the cuDNN framework \cite{DBLP:journals/corr/ChetlurWVCTCS14}.

\section{V-trace}
\label{sec:v-trace}

Off-policy learning is important in the decoupled distributed actor-learner architecture because of the lag between when actions are generated by the actors and when the learner estimates the gradient. To this end, we introduce a novel off-policy actor-critic algorithm for the learner, called V-trace. 

First, let us introduce some notations. We consider the problem of discounted infinite-horizon RL in Markov Decision Processes (MDP), see \cite{Puterman:1994,sutton-barto98} where the goal is to find a policy $\pi$ that maximises the expected sum of future discounted rewards:
$V^{\pi}(x) \eqdef\E_{\pi}\big[\sum_{t\geq 0} \gamma^t r_t\big]$, where $\gamma\in[0,1)$ is the discount factor, $r_t = r(x_t, a_t)$ is the reward at time $t$, $x_t$ is the state at time $t$ (initialised in $x_0=x$) and $a_t\sim\pi(\cdot|x_t)$ is the action generated by following some policy $\pi$. 

The goal of an off-policy RL algorithm is to use trajectories generated by some policy $\mu$, called the {\em behaviour policy}, to learn the value function $V^{\pi}$ of another policy $\pi$ (possibly different from $\mu$), called the {\em target policy}. 

\subsection{V-trace target}
Consider a trajectory $\left(x_t, a_t, r_t\right)_{t=s}^{t=s+n}$ generated by the actor following some policy $\mu$. We define the $n$-steps V-trace target for $V(x_s)$, our value approximation at state $x_s$, as:
\beqa
v_s &\eqdef& \textstyle{V(x_s) + \sum_{t=s}^{s+n-1} \gamma^{t-s} \Big( \prod_{i=s}^{t-1} c_i \Big ) \delta_t V}, \label{eq:target.off}
\eeqa
where $\delta_t V \eqdef \rho_t \big(r_t+\gamma V(x_{t+1})-V(x_t)\big)$  is a temporal difference for $V$, and  $\rho_t\eqdef\min\big(\bar\rho, \frac{\pi(a_t|x_t)}{\mu(a_t|x_t)}\big)$ and $c_i\eqdef\min\big(\bar c, \frac{\pi(a_i|x_i)}{\mu(a_i|x_i)}\big)$ are truncated importance sampling (IS) weights (we make use of the notation $\prod_{i=s}^{t-1} c_i=1$ for $s=t$). In addition we assume that the truncation levels are such that $\bar \rho\geq \bar c$.

Notice that in the on-policy case (when $\pi=\mu$), and assuming that $\bar c \geq 1$, then all $c_i=1$ and $\rho_t=1$, thus   (\ref{eq:target.off}) rewrites
\begin{align}
v_s &= V(x_s) + \textstyle{\sum_{t=s}^{s+n-1}} \gamma^{t-s} \big(r_t+\gamma V(x_{t+1})-V(x_t)\big) \notag \\
&= 
\label{eq:target.on}
\textstyle{\sum_{t=s}^{s+n-1}} \gamma^{t-s} r_t + \gamma^{n} V(x_{s+n}),
\end{align}
which is the on-policy $n$-steps Bellman target. Thus in the on-policy case, V-trace reduces to the on-policy $n$-steps Bellman update. This property (which Retrace \cite{munos2016safe} does not have) allows one to use the same algorithm for off- and on-policy data. 
 
Notice that the (truncated) IS weights $c_i$ and $\rho_t$ play different roles.  The weight $\rho_t$ appears in the definition of the temporal difference $\delta_t V$ and defines the fixed point of this update rule. In a tabular case, where functions can be perfectly represented, the fixed point of this update (i.e., when $V(x_s)=v_s$ for all states), characterised by $\delta_t V$ being equal to zero in expectation (under $\mu$), is the value function $V^{\pi_{\bar \rho}}$ of some policy $\pi_{\bar \rho}$, defined by
 \beq\label{eq:pi_rho}
 \pi_{\bar \rho}(a|x)\eqdef \frac{\min \big(\bar \rho \mu(a|x),\pi(a|x)\big)}{\sum_{b\in A}\min \big(\bar \rho \mu(b|x),\pi(b|x)\big)},
 \eeq
  (see the analysis in Appendix~\ref{sec:analysis}
).
So when $\bar \rho$ is infinite (i.e.~no truncation of $\rho_t$), then this is the value function $V^{\pi}$ of the target policy.  However if we choose a truncation level $\bar \rho<\infty$, our fixed point is the value function $V^{\pi_{\bar \rho}}$ of a policy $\pi_{\bar\rho}$ which is somewhere between $\mu$ and $\pi$. At the limit when $\bar \rho$ is close to zero, we obtain the value function of the behaviour policy $V^{\mu}$. In Appendix~\ref{sec:analysis}
we prove the contraction of a related V-trace operator and the convergence of the corresponding online V-trace algorithm.

The weights $c_i$ are similar to the ``trace cutting" coefficients in Retrace. Their product $c_s\dots c_{t-1}$ measures how much a temporal difference $\delta_tV$ observed at time $t$ impacts the update of the value function at a previous time $s$. The more dissimilar $\pi$ and $\mu$ are (the more off-policy we are), the larger the variance of this product. We use the truncation level $\bar c$ as a variance reduction technique. However notice that this truncation does not impact the solution to which we converge (which is characterised by $\bar \rho$ only).

Thus we see that the truncation levels $\bar c$ and $\bar \rho$ represent different features of the algorithm:  $\bar \rho$ impacts the nature of the value function we converge to, whereas $\bar c$ impacts the speed at which we converge to this function.
 
\begin{remark}
V-trace targets can be computed recursively:
$$v_s = V(x_s) + \delta_sV + \gamma c_s\big( v_{s+1} - V(x_{s+1})\big).$$
\end{remark}
\begin{remark}
Like in Retrace($\lambda$), we can also consider an additional discounting parameter $\lambda\in [0,1]$ in the definition of V-trace by setting $c_i=\lambda\min\big(\bar c, \frac{\pi(a_i|x_i)}{\mu(a_i|x_i)}\big)$. In the on-policy case, when $n=\infty$, V-trace then reduces to TD($\lambda$).
\end{remark}

\subsection{Actor-Critic algorithm}

\subsubsection*{Policy gradient}
In the on-policy case, the gradient of the value function $V^{\mu}(x_0)$ with respect to some parameter of the policy $\mu$ is
$$\nabla V^{\mu}(x_0) = \E_{\mu}\Big[ \textstyle{\sum_{s\geq 0}}\gamma^s \nabla\log\mu(a_s|x_s) Q^{\mu}(x_s, a_s)\Big],$$
where $Q^{\mu}(x_s, a_s)\eqdef \E_{\mu} \big[ \sum_{t\geq s}\gamma^{t-s} r_t |x_s,a_s \big]$ is the state-action value of policy $\mu$ at $(x_s,a_s)$. This is usually implemented by a stochastic gradient ascent that updates the policy parameters in the direction of 
$\E_{a_s\sim \mu(\cdot|x_s)}\Big[ \nabla\log\mu(a_s|x_s) q_s \big| x_s\Big]$, 
where $q_s$ is an estimate of $Q^{\mu}(x_s,a_s)$, and averaged over the set of states $x_s$ that are visited under some behaviour policy $\mu$.

Now in the off-policy setting that we consider, we can use an IS weight between the policy being evaluated $\pi_{\bar\rho}$ and the behaviour policy $\mu$, to update our policy parameter in the direction of
\beq\label{eq:off-policy.PG}
\E_{a_s\sim \mu(\cdot|x_s)}\Big[ \frac{\pi_{\bar \rho}(a_s|x_s)}{\mu(a_s|x_s)} \nabla\log\pi_{\bar \rho}(a_s|x_s) q_s \big| x_s\Big]
\eeq
where $q_s \eqdef r_s + \gamma v_{s+1}$ is an estimate of  $Q^{\pi_{\bar \rho}}(x_s, a_s)$ built from the V-trace estimate $v_{s+1}$ at the next state $x_{s+1}$.
The reason why we use $q_s$ instead of $v_s$ as the target for our Q-value $Q^{\pi_{\bar \rho}}(x_s,a_s)$ is that, assuming our value estimate is correct at all states, i.e.~$V=V^{\pi_{\bar \rho}}$, then we have $\E [q_s|x_s,a_s] = Q^{\pi_{\bar \rho}}(x_{s},a_s)$ (whereas we do not have this property if we choose $q_t = v_t$). See Appendix~\ref{sec:analysis}
for analysis and Appendix~\ref{sec:appendix_q_estimation}
for a comparison of different ways to estimate $q_s$.

In order to reduce the variance of the policy gradient estimate~\eqref{eq:off-policy.PG}, we usually subtract from $q_s$ a state-dependent baseline, such as the current value approximation $V(x_s)$.

Finally notice that \eqref{eq:off-policy.PG} estimates the policy gradient for $\pi_{\bar \rho}$ which is the policy evaluated by the V-trace algorithm when using a truncation level $\bar \rho$. However assuming the bias $V^{\pi_{\bar\rho}}-V^{\pi}$ is small (e.g.~if $\bar\rho$ is large enough) then we can expect $q_s$ to provide us with a good estimate of $Q^{\pi}(x_s,a_s)$. Taking into account these remarks, we derive the following canonical V-trace actor-critic algorithm.

\subsubsection*{V-trace actor-critic algorithm}
Consider a parametric representation $V_{\theta}$ of the value function and the current policy $\pi_{\omega}$. Trajectories have been generated by actors following some behaviour policy $\mu$. The V-trace targets $v_s$ are defined by \eqref{eq:target.off}. At training time $s$, the value parameters $\theta$ are updated by gradient descent on the $l2$ loss to the target $v_s$, i.e., in the direction of
$$\big( v_s - V_\theta(x_s)\big) \nabla_\theta V_\theta(x_s),$$
and the policy parameters $\omega$ in the direction of the policy gradient:
$$ \rho_s \nabla_\omega\log\pi_\omega(a_s|x_s) \big( r_s+\gamma v_{s+1} - V_\theta(x_s)\big).
$$
In order to prevent premature convergence we may add an entropy bonus, like in A3C, along the direction
$$-\nabla_\omega \sum_a \pi_\omega(a|x_s) \log\pi_\omega(a|x_s).$$
The overall update is obtained by summing these three gradients rescaled by appropriate coefficients, which are hyperparameters of the algorithm.

\section{Experiments}
\label{sec:experiments}
We investigate the performance of IMPALA under multiple settings.
For data efficiency, computational performance and effectiveness of the off-policy correction we look at the learning behaviour of IMPALA agents trained on individual tasks.
For multi-task learning we train agents---each with one set of weights for all tasks---on a newly introduced collection of 30 DeepMind Lab tasks and on all 57 games of the Atari Learning Environment \cite{bellemare13arcade}.

\begin{figure}
    \includegraphics[width=.42\columnwidth]{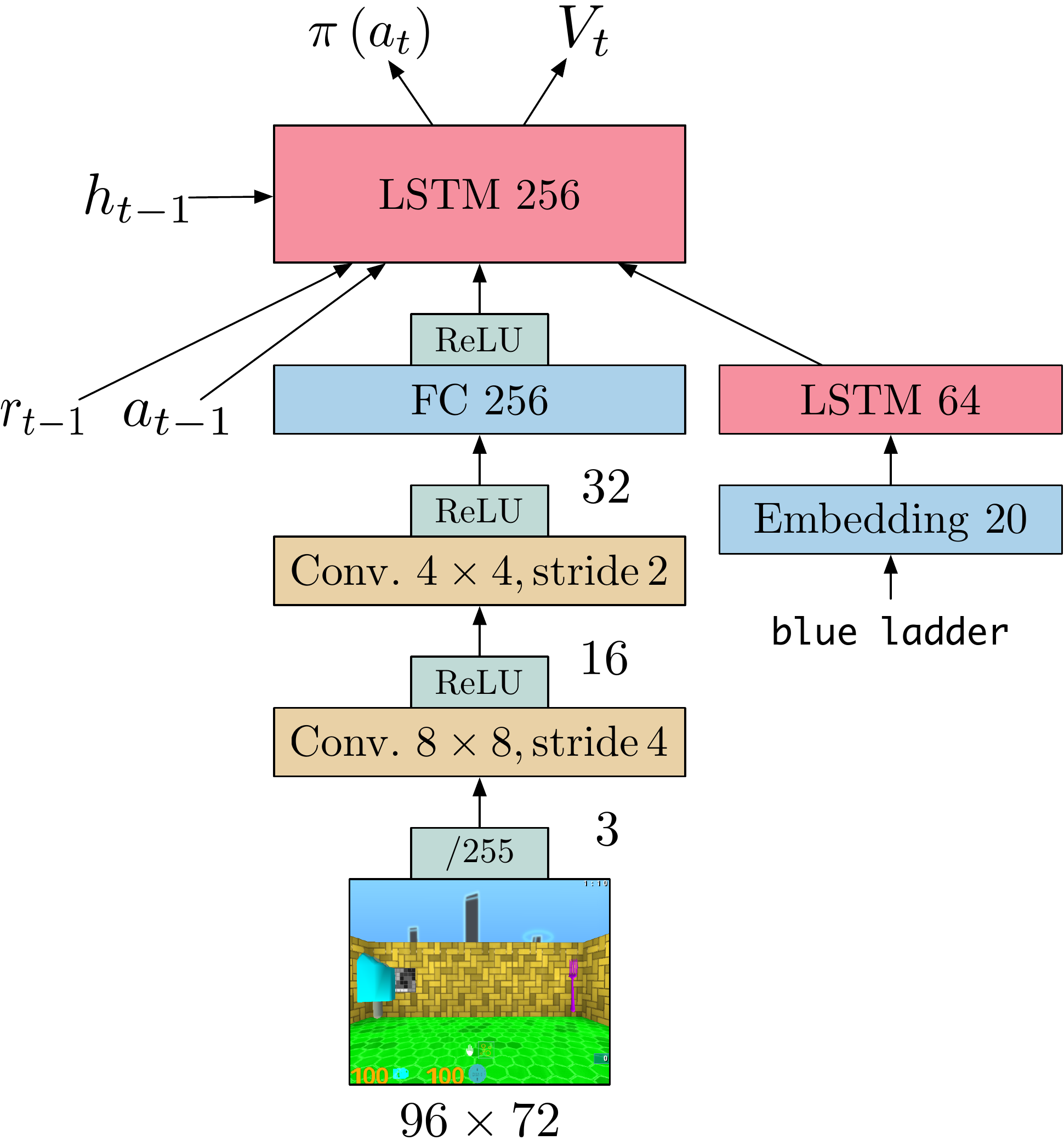}
    \hfill
    \includegraphics[width=.56\columnwidth]{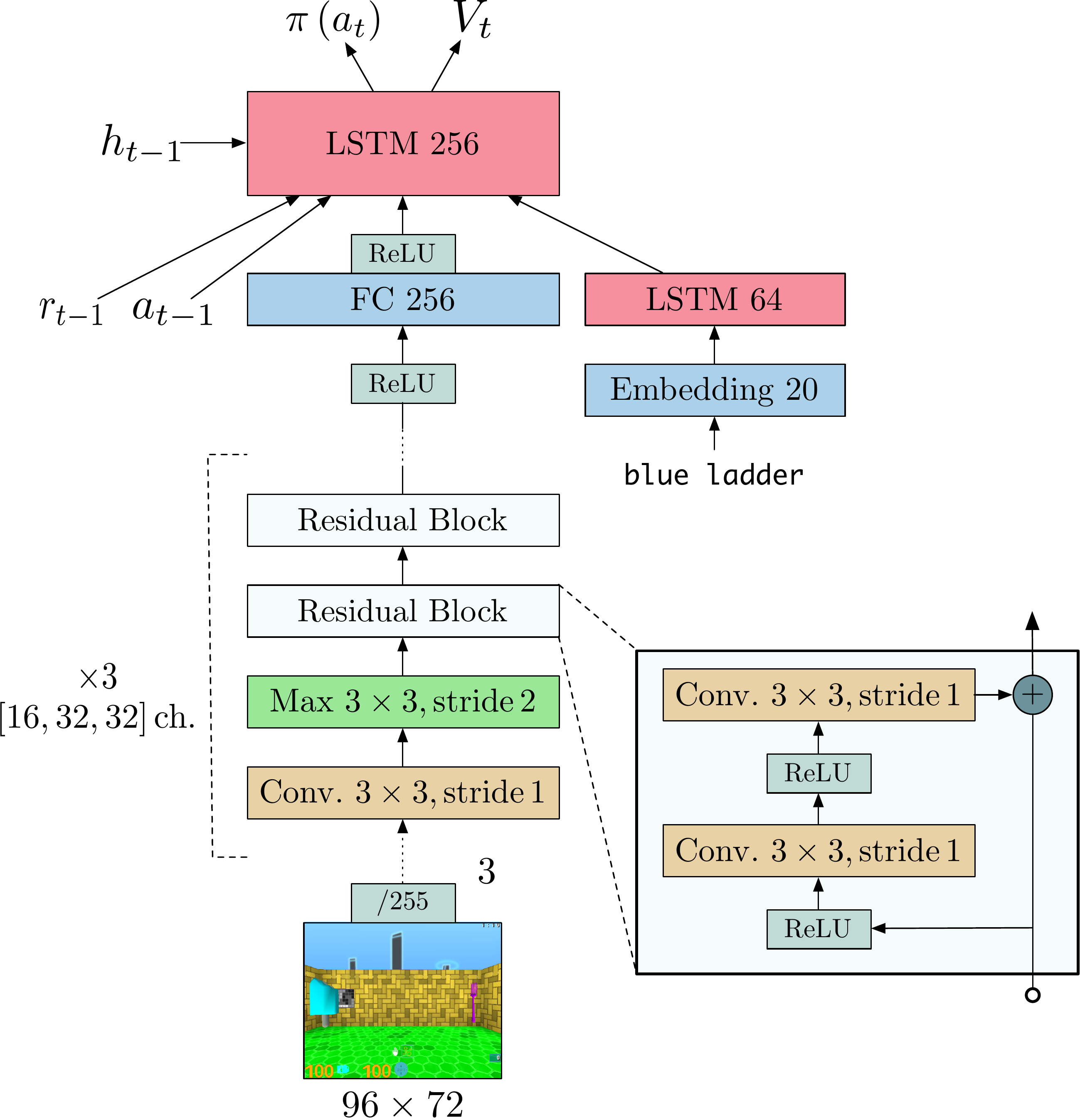}
    \caption{Model Architectures. {\bf Left:} Small architecture, $2$ convolutional layers and $1.2$ million parameters. 
    {\bf Right:} Large architecture, $15$ convolutional layers and $1.6$ million parameters.}
    \label{model_single}
    \label{model_dmlab30}
\end{figure}

For all the experiments we have used two different model architectures: a shallow model similar to \cite{A3C2016} with an LSTM before the policy and value (shown in Figure~\ref{model_single} (left)) and a deeper residual model \cite{he2016identity} (shown in Figure~\ref{model_dmlab30} (right)). For tasks with a language channel we used an LSTM with text embeddings as input.

\subsection{Computational Performance}

\begin{table}[t]
\small
  \begin{center}
  \begin{threeparttable}
  \begin{tabular}{l@{\hspace{.22cm}}c@{\hspace{.22cm}}c@{\hspace{.22cm}}c@{\hspace{.22cm}}c@{\hspace{.22cm}}}
    \toprule
    \textbf{Architecture} & \textbf{CPUs} &  \textbf{GPUs\tnote{1}}  & \multicolumn{2}{c}{\textbf{FPS\tnote{2}}} \\
    \midrule
    \textbf{Single-Machine} &               &                 & \scriptsize Task 1   & \scriptsize Task 2 \\
    \midrule
      A3C \scriptsize 32 workers                                     & 64  & 0 & 6.5K  & 9K   \\
      Batched A2C\scriptsize\ (sync step)         & 48  & 0 & 9K & 5K   \\
      Batched A2C\scriptsize\ (sync step)         & 48  & 1 & 13K & 5.5K   \\
      Batched A2C\scriptsize\ (sync traj.)         & 48  & 0 & 16K & 17.5K   \\
      Batched A2C \scriptsize (dyn. batch)              & 48  & 1 & 16K & 13K  \\
      IMPALA \scriptsize 48 actors             & 48  & 0 & 17K & 20.5K  \\
      IMPALA \scriptsize (dyn. batch) 48 actors\tnote{3}             & 48  & 1 & 21K & 24K  \\
      \midrule
      \textbf{Distributed} \\
      \midrule
      A3C                         & 200 & 0 & 46K & 50K  \\
      IMPALA \scriptsize                        & 150 & 1 & \multicolumn{2}{c}{80K}  \\
      IMPALA \scriptsize (optimised)            & 375 & 1 & \multicolumn{2}{c}{200K} \\
      IMPALA \scriptsize (optimised) batch 128           & 500 & 1 & \multicolumn{2}{c}{250K} \\
      \bottomrule
  \end{tabular}
  \tiny $^1$ Nvidia P100 $^2$  In frames/sec (4 times the agent steps due to action repeat). $^3$ Limited by amount of rendering possible on a single machine.
  \end{threeparttable}
  \end{center}
\caption{Throughput on \texttt{seekavoid\_arena\_01} (task 1) and \texttt{rooms\_keys\_doors\_puzzle} (task 2) with the shallow model in Figure~\ref{model_single}. The latter has variable length episodes and slow restarts. Batched A2C and IMPALA use batch size 32 if not otherwise mentioned.}
\label{speed_single}
\end{table}

High throughput, computational efficiency and scalability are among the main design goals of IMPALA.
To demonstrate that IMPALA outperforms current algorithms in these metrics we compare A3C~\cite{A3C2016}, batched A2C variations and IMPALA variants with various optimisations. For single-machine experiments using GPUs, we use dynamic batching in the forward pass to avoid several batch size 1 forward passes. Our dynamic batching module is implemented by specialised TensorFlow operations but is conceptually similar to the queues used in GA3C.
Table~\ref{speed_single} details the results for single-machine and multi-machine versions with the shallow model from Figure~\ref{model_single}.
In the single-machine case, IMPALA achieves the highest performance on both tasks, ahead of all batched A2C variants and ahead of A3C.
However, the distributed, multi-machine setup is where IMPALA can really demonstrate its scalability. With the optimisations from Section~\ref{sec:efficiency_optimizations} to speed up the GPU-based learner, the IMPALA agent achieves a throughput rate of 250,000 frames/sec or $21$ billion frames/day.
Note, to reduce the number of actors needed per learner, one can use auxiliary losses, data from experience replay or other expensive learner-only computation.

\subsection{Single-Task Training}

To investigate IMPALA's learning dynamics, we employ the single-task scenario where we train agents individually on 5 different DeepMind Lab tasks. The task set consists of a planning task, two maze navigation tasks, a laser tag task with scripted bots and a simple fruit collection task.

We perform hyperparameter sweeps over the weighting of \textit{entropy regularisation}, the \textit{learning rate} and the \textit{RMSProp epsilon}. For each experiment we use an identical set of 24 pre-sampled hyperparameter combinations from the ranges in Appendix~\ref{tab:hyperparameter_ranges}
. The other hyperparameters were fixed to values specified in Appendix~\ref{tab:fixed_model_hyperparameters}
.

\subsubsection{Convergence and Stability}

\begin{figure*}
    \includegraphics[width=1.0\textwidth]{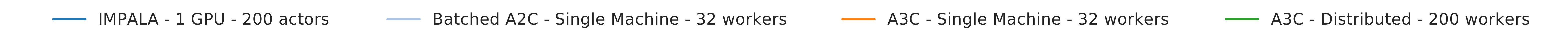}
    \includegraphics[width=0.19\textwidth]{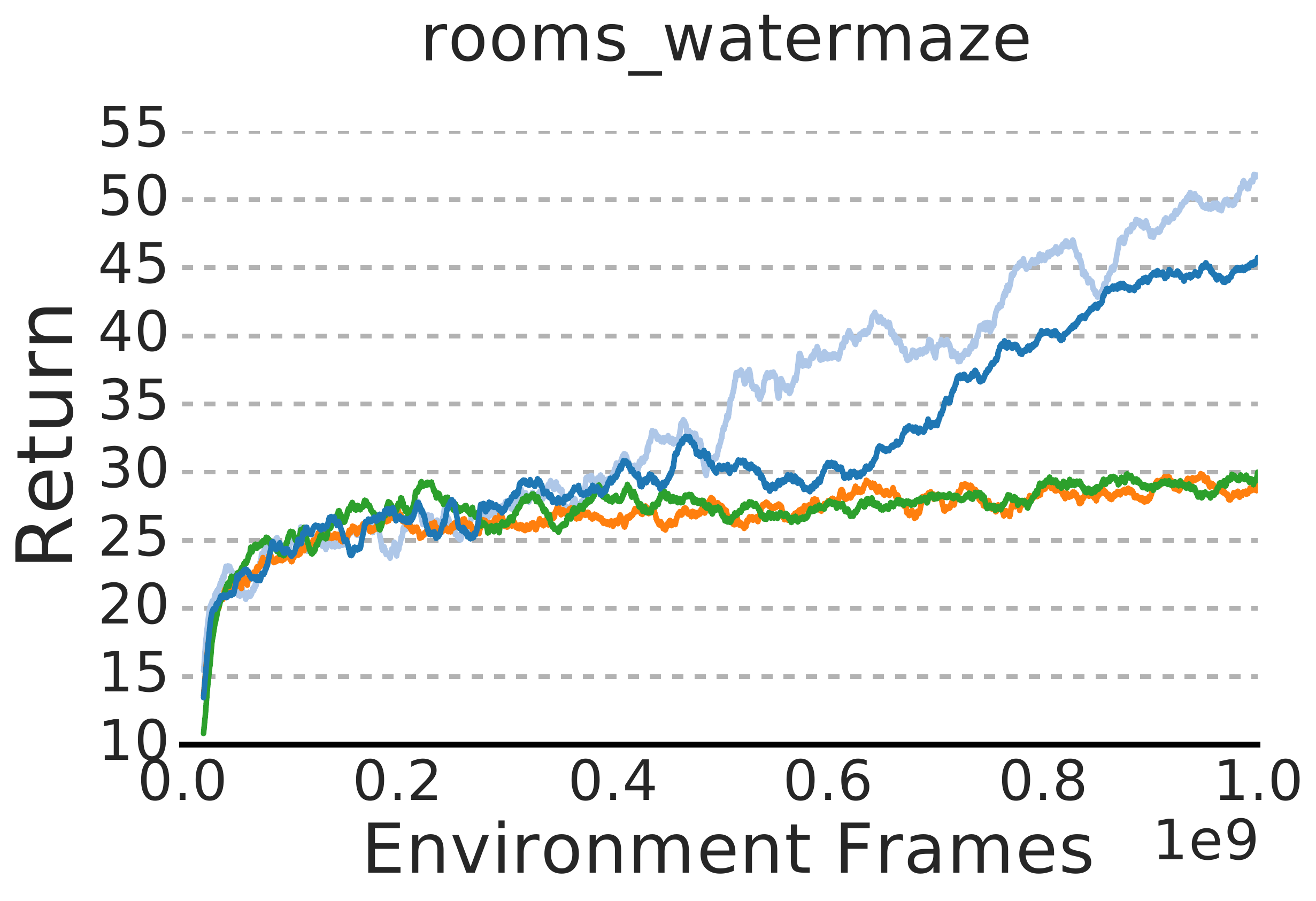}
    \includegraphics[width=0.19\textwidth]{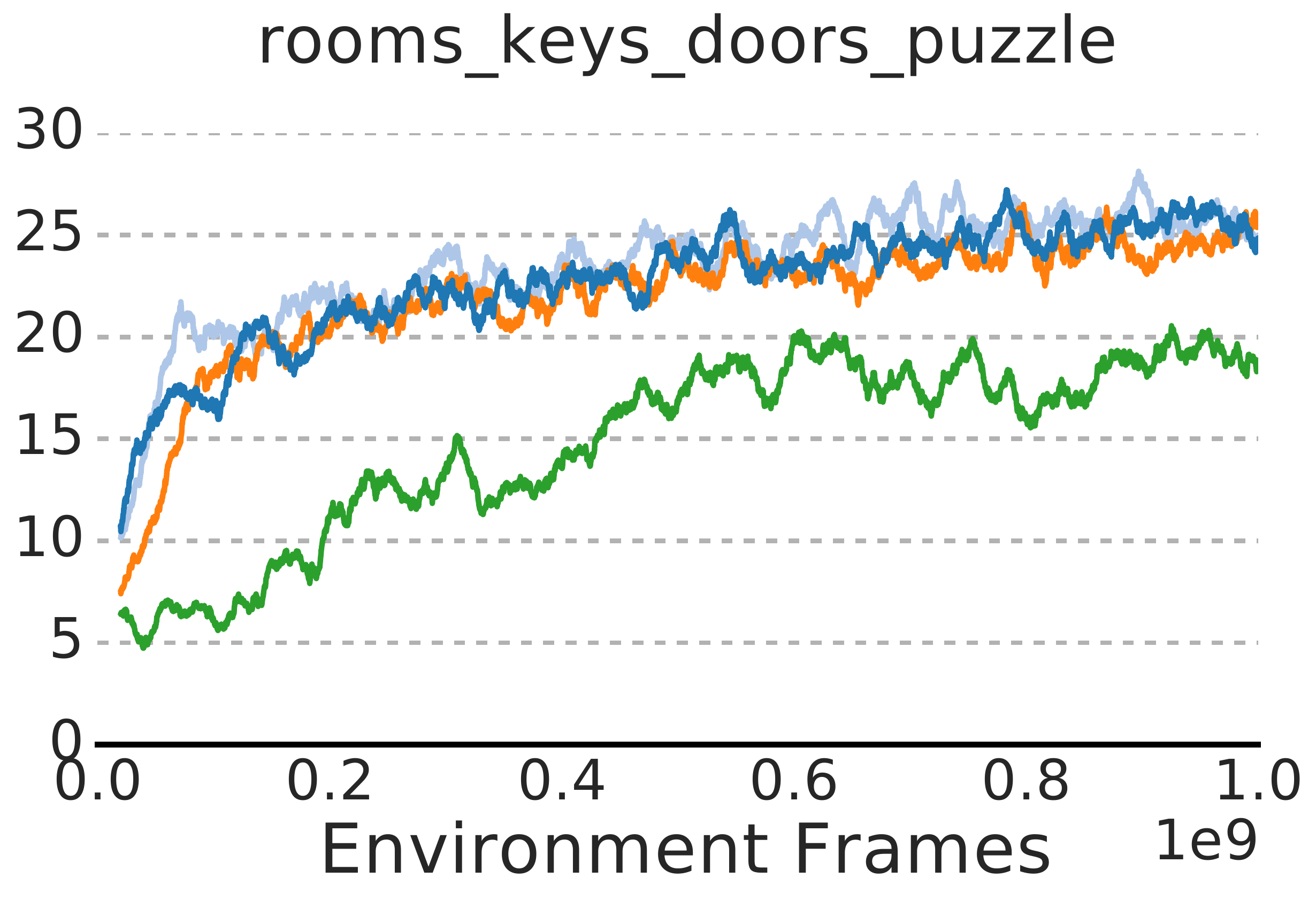}
    \includegraphics[width=0.19\textwidth]{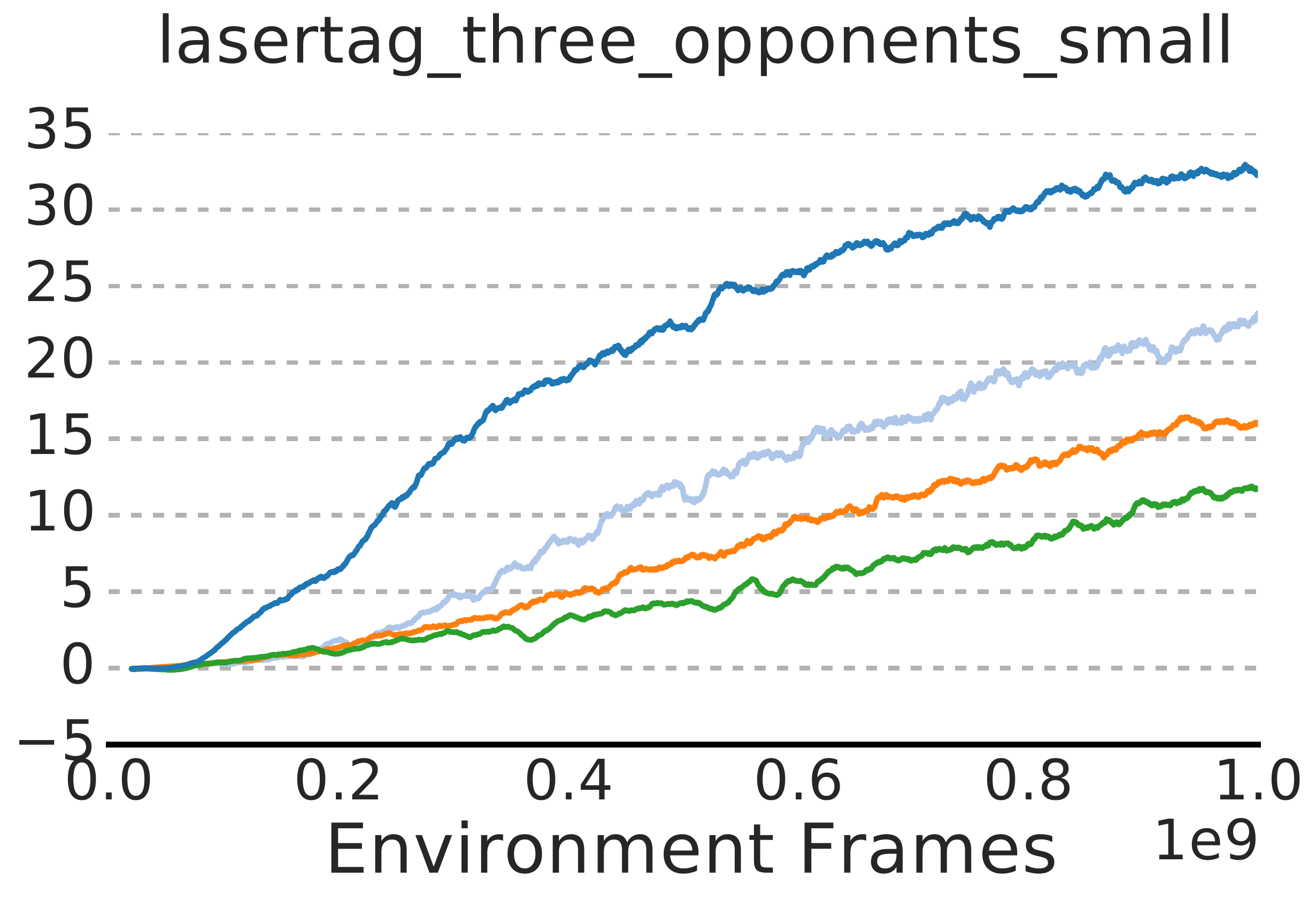}
    \includegraphics[width=0.19\textwidth]{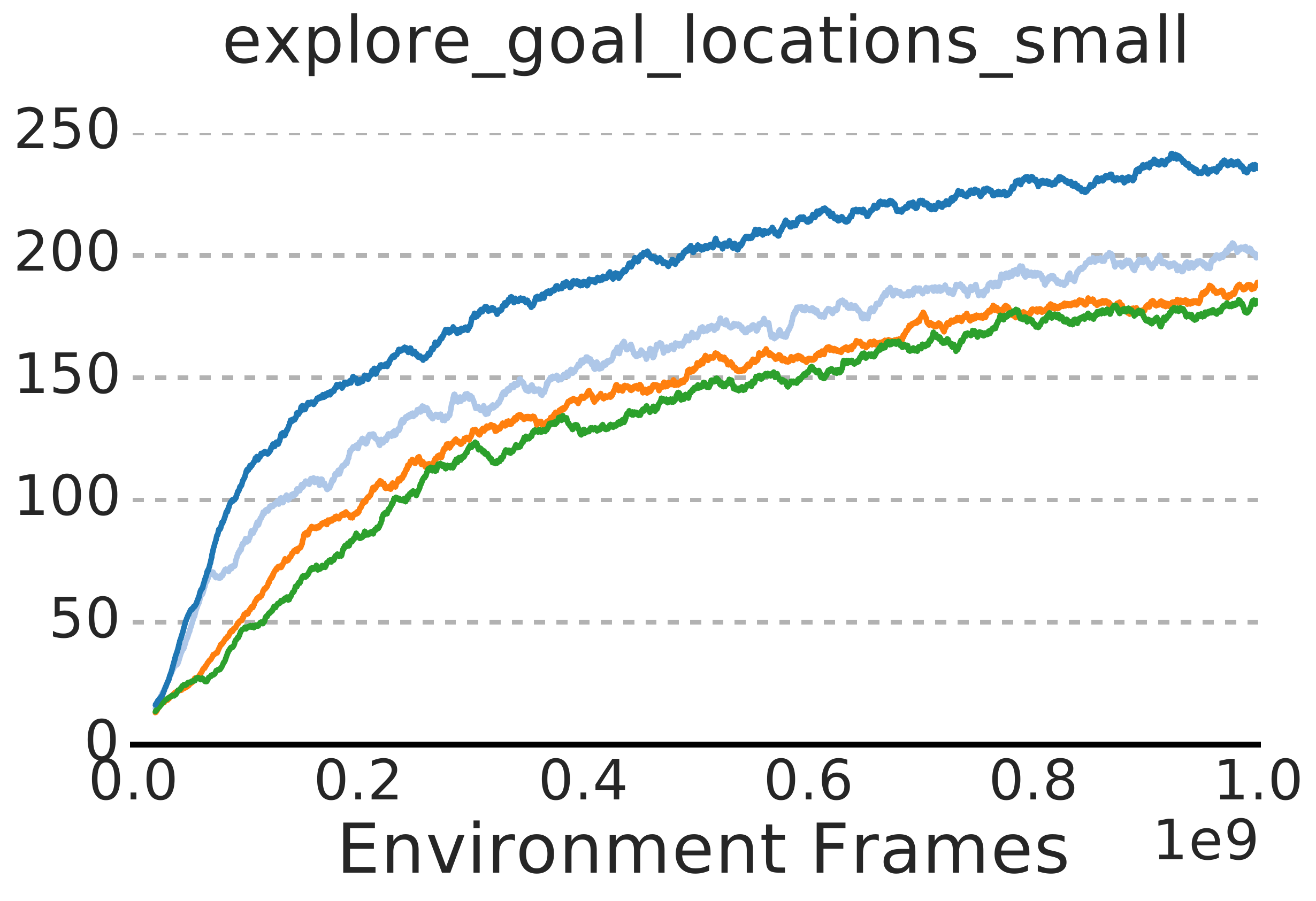}
    \includegraphics[width=0.19\textwidth]{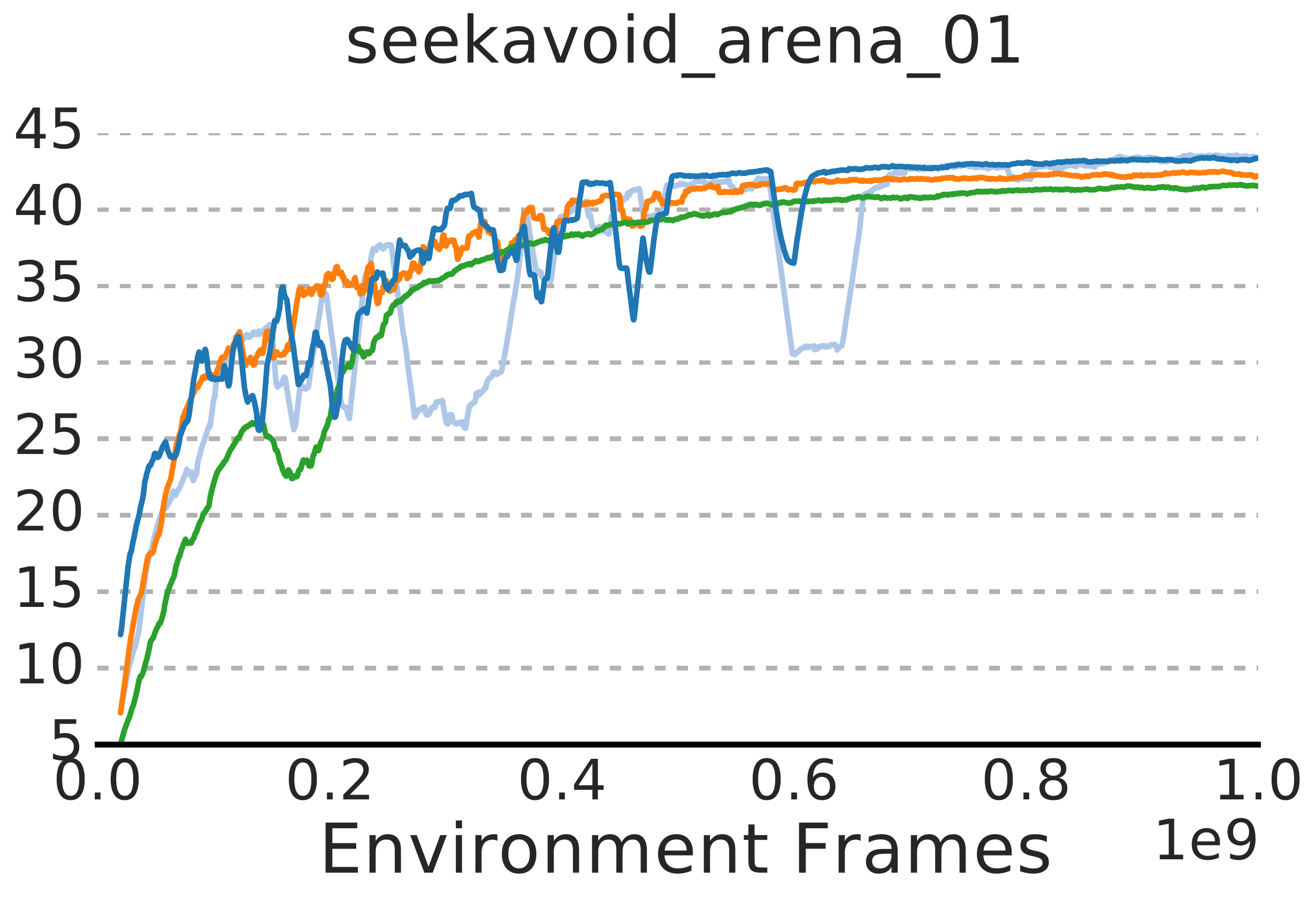}
    \\
    \includegraphics[width=0.19\textwidth]{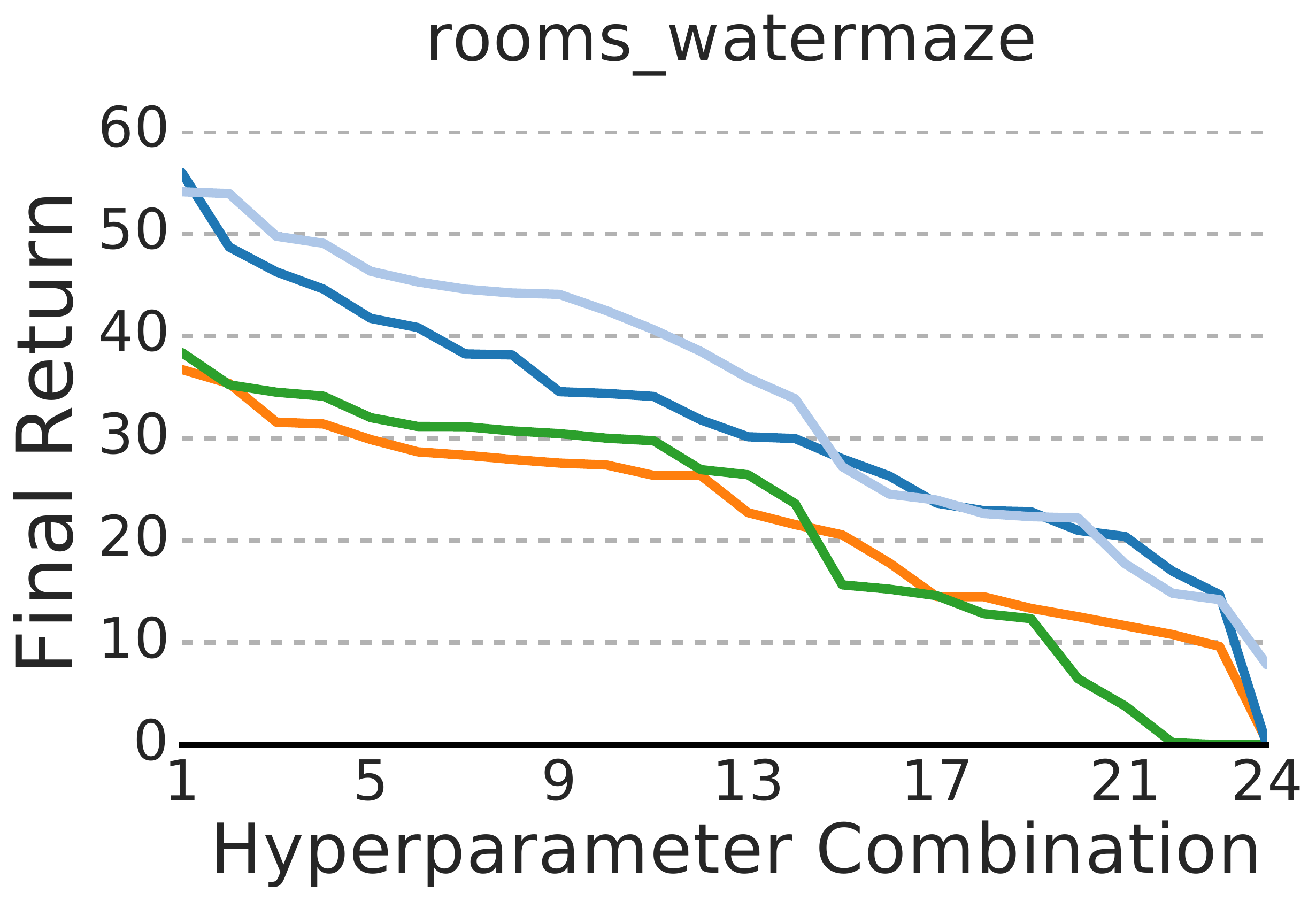}
    \includegraphics[width=0.19\textwidth]{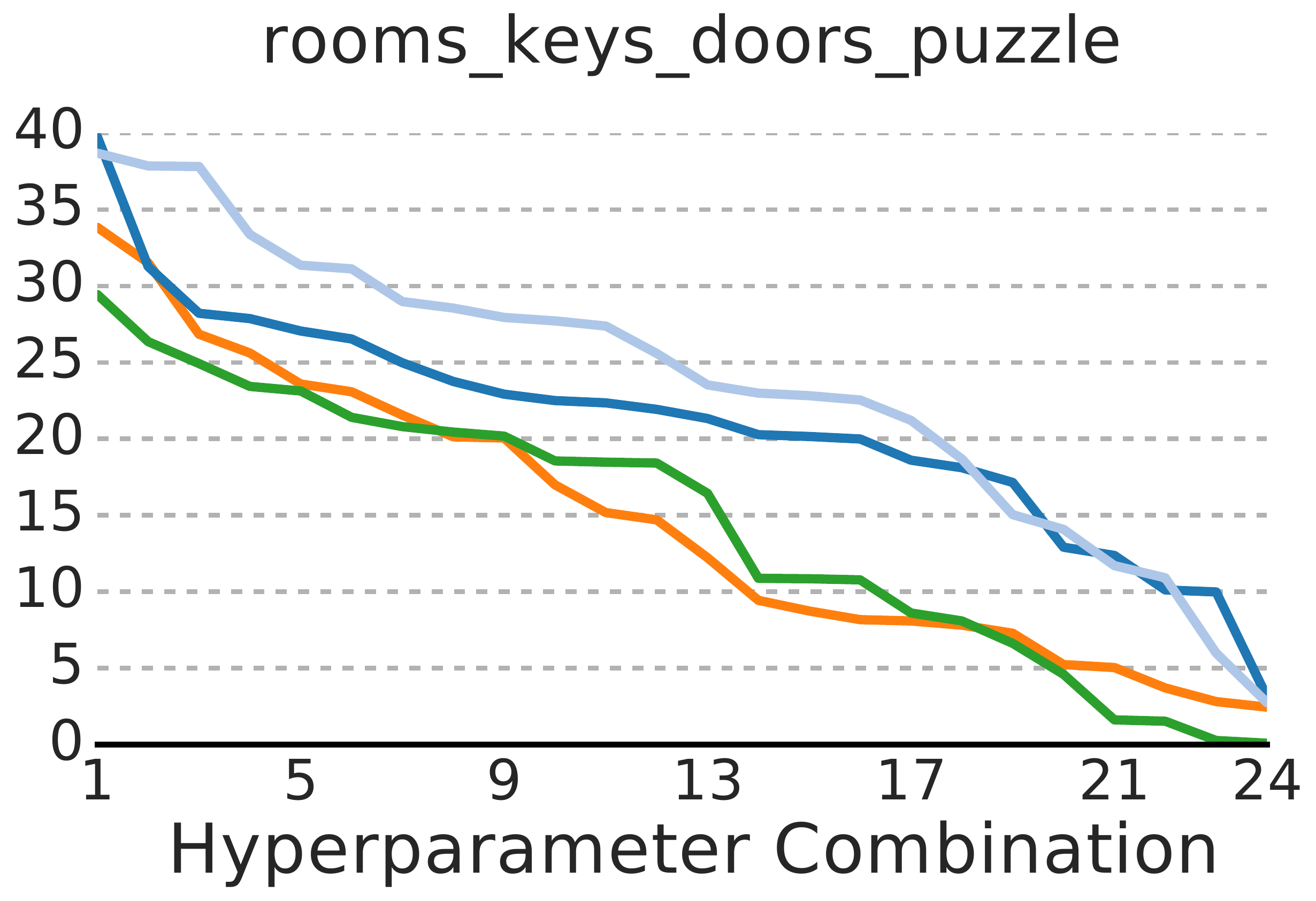}
    \includegraphics[width=0.19\textwidth]{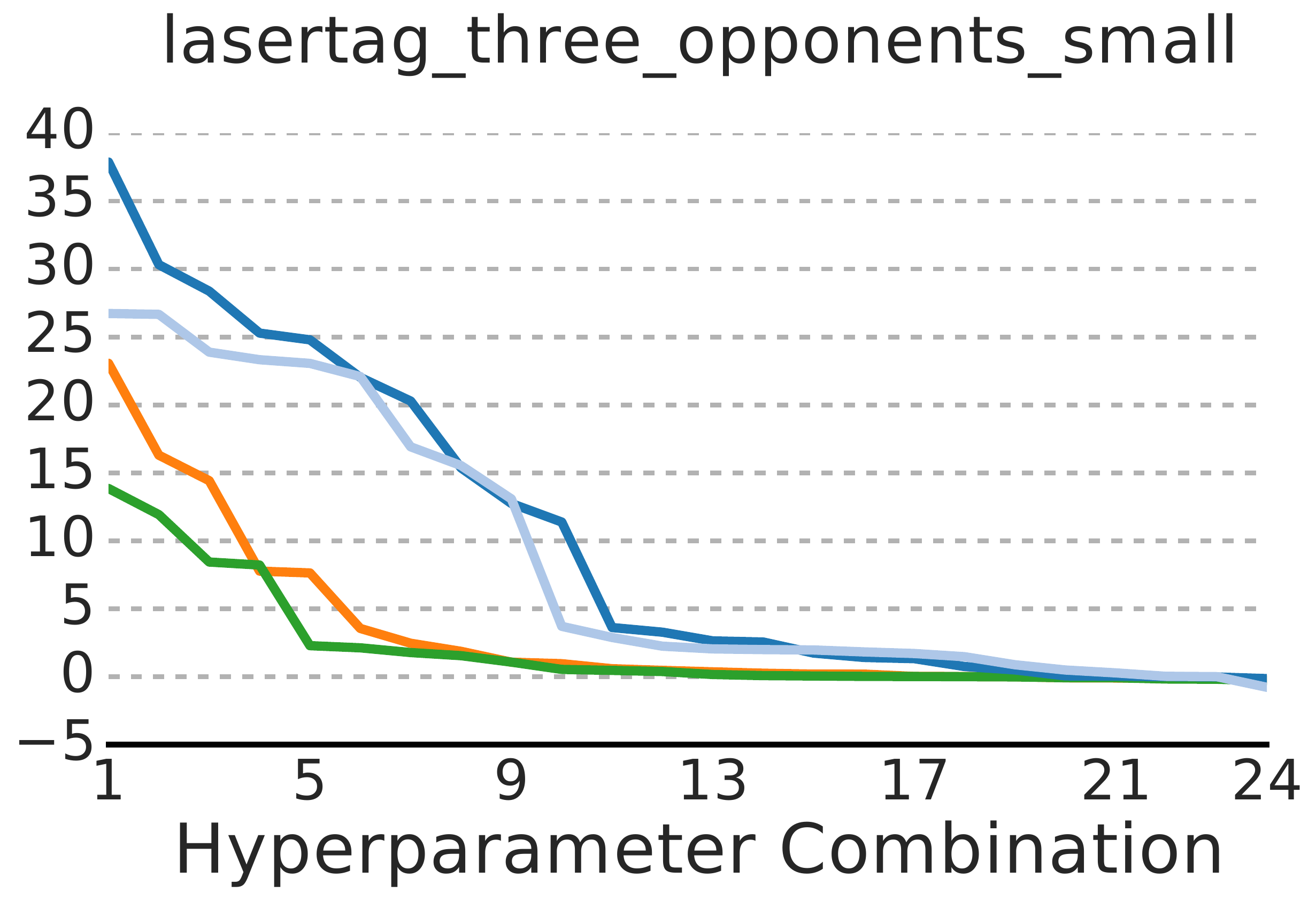}
    \includegraphics[width=0.19\textwidth]{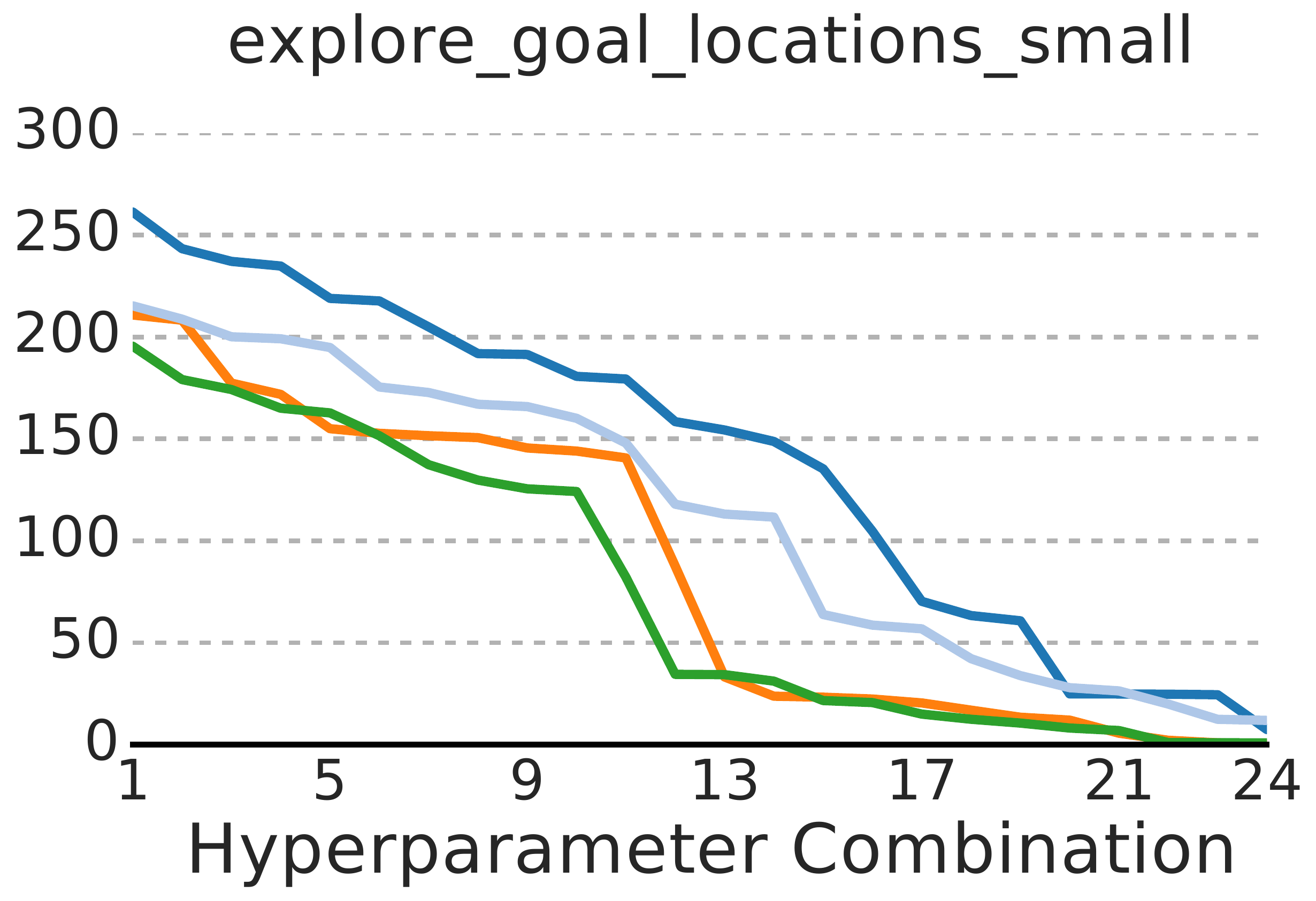}
    \includegraphics[width=0.19\textwidth]{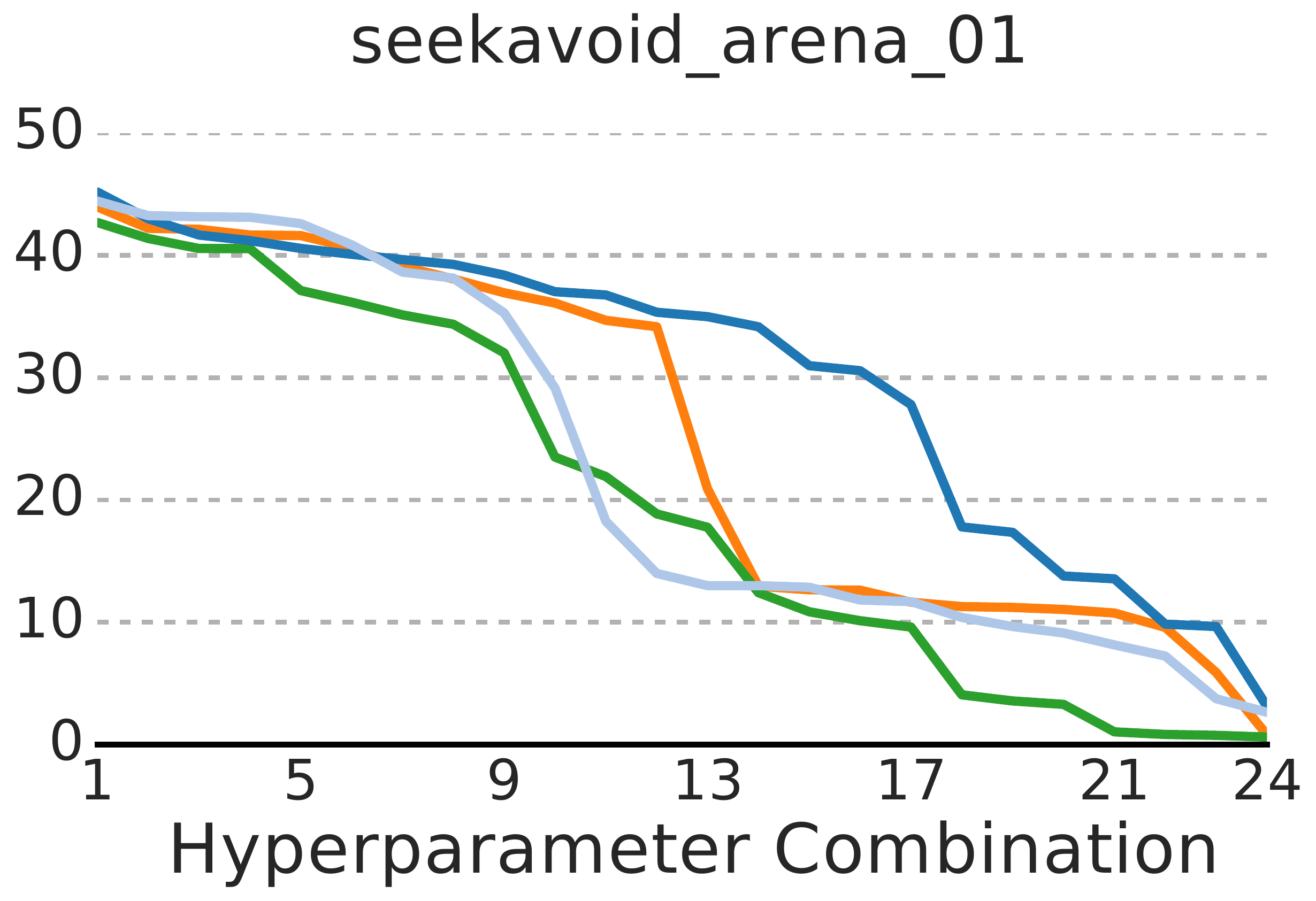}
    
    \caption{\textbf{Top Row:} Single task training on 5 DeepMind Lab tasks. Each curve is the mean of the best 3 runs based on final return. IMPALA achieves better performance than A3C. \textbf{Bottom Row:} Stability across hyperparameter combinations sorted by the final performance across different hyperparameter combinations. IMPALA is consistently more stable than A3C.}
    \label{single_level_learning_curves}
    \label{stability_vs_a3c}
\end{figure*}

Figure~\ref{single_level_learning_curves} shows a comparison between IMPALA, A3C and batched A2C with the shallow model in Figure~\ref{model_single}.
In all of the 5 tasks, either batched A2C or IMPALA reach the best final average return and in all tasks but \texttt{seekavoid\_arena\_01} they are ahead of A3C throughout the entire course of training. IMPALA outperforms the synchronous batched A2C on 2 out of 5 tasks while achieving much higher throughput (see Table~\ref{speed_single}).
We hypothesise that this behaviour could stem from the V-trace off-policy correction acting similarly to generalised advantage estimation \cite{schulman2015high} and asynchronous data collection yielding more diverse batches of experience.

In addition to reaching better final performance, IMPALA is also more robust to the choice of hyperparameters than A3C. Figure~\ref{stability_vs_a3c} compares the final performance of the aforementioned methods across different hyperparameter combinations, sorted by average final return from high to low. Note that IMPALA achieves higher scores over a larger number of combinations than A3C.

\subsubsection{V-trace Analysis}
\label{sec:vtrace_variants}

\begin{table}[t]
\small
  \begin{center}
  \begin{threeparttable}
  \begin{tabular}{l@{\hspace{.22cm}}c@{\hspace{.22cm}}c@{\hspace{.22cm}}c@{\hspace{.22cm}}c@{\hspace{.22cm}}c@{\hspace{.22cm}}}
  \toprule
  & \scriptsize Task 1 & \scriptsize Task 2 & \scriptsize Task 3 & \scriptsize Task 4 & \scriptsize Task 5 \\
  \midrule
  \textbf{Without Replay} \\
  \midrule
  V-trace        &         46.8    &         32.9      & \textbf{31.3}              & \textbf{229.2}  & \textbf{43.8}   \\   
  1-Step         & \textbf{51.8}   & \textbf{35.9}     &         25.4               &         215.8   &         43.7    \\   
  $\eps$-correction           &         44.2    &         27.3      &         4.3                &         107.7   &         41.5    \\   
  No-correction  &         40.3    &         29.1      &         5.0                &         94.9    &         16.1    \\
  \midrule
  \textbf{With Replay} \\
  \midrule
  V-trace        &         47.1    & \textbf{35.8}     & \textbf{34.5}              & \textbf{250.8}  & \textbf{46.9} \\
  1-Step         & \textbf{54.7}   &         34.4      &         26.4               &         204.8   &         41.6  \\
  $\eps$-correction           &         30.4    &         30.2      &         3.9                &         101.5   &         37.6  \\
  No-correction  &         35.0    &         21.1      &         2.8                &         85.0    &         11.2  \\
  \bottomrule
  \end{tabular}
  \begin{flushleft}
  \tiny Tasks: \texttt{rooms\_watermaze}, \texttt{rooms\_keys\_doors\_puzzle}, \texttt{lasertag\_three\_opponents\_small}, \texttt{explore\_goal\_locations\_small}, \texttt{seekavoid\_arena\_01}
  \end{flushleft}
  \end{threeparttable}
  \end{center}
  \caption{Average final return over 3 best hyperparameters for different off-policy correction methods on 5 DeepMind Lab tasks. When the lag in policy is negligible both V-trace and 1-step importance sampling perform similarly well and better than $\epsilon$-correction/No-correction. However, when the lag increases due to use of experience replay, V-trace performs better than all other methods in $4$ out $5$ tasks.}
  \label{tab:vtrace_analysis}
\end{table}

To analyse V-trace we investigate four different algorithms:\\
    \textbf{1. No-correction} - No off-policy correction.\\
    \textbf{2.\hspace{.1cm}$\epsilon$-correction} - Add a small value ($\epsilon = 1e\mbox{-}6$) during gradient calculation to prevent $\log \pi(a)$ from becoming very small and leading to numerical instabilities, similar to~\cite{babaeizadeh2016ga3c}.\\
    \textbf{3. 1-step importance sampling} - No off-policy correction when optimising $V(x)$. For the policy gradient, multiply the advantage at each time step by the corresponding importance weight. This variant is similar to V-trace without ``traces" and is included to investigate the importance of ``traces" in V-trace.\\
    \textbf{4. V-trace} as described in Section~\ref{sec:v-trace}.

For V-trace and 1-step importance sampling we clip each importance weight $\rho_t$ and $c_t$ at $1$ (i.e. $\bar c = \bar \rho = 1)$. This reduces the variance of the gradient estimate but introduces a bias. Out of $\bar\rho\in [1, 10, 100]$ we found that $\bar\rho = 1$ worked best.

We evaluate all algorithms on the set of 5 DeepMind Lab tasks from the previous section. We also add an experience replay buffer on the learner to increase the off-policy gap between $\pi$ and $\mu$. In the experience replay experiments we draw 50\% of the items in each batch uniformly at random from the replay buffer.
Table~\ref{tab:vtrace_analysis} shows the final performance for each algorithm with and without replay respectively. In the no replay setting, V-trace performs best on 3 out of 5 tasks, followed by 1-step importance sampling, $\eps$-correction and No-correction.
Although 1-step importance sampling performs similarly to V-trace in the no-replay setting, the gap widens on 4 out 5 tasks when using experience replay. This suggests that the cruder 1-step importance sampling approximation becomes insufficient as the target and behaviour policies deviate from each other more strongly. Also note that V-trace is the only variant that consistently benefits from adding experience replay.
$\eps$-correction improves significantly over No-correction on two tasks but lies far behind the importance-sampling based methods, particularly in the more off-policy setting with experience replay. Figure~\ref{fig:policy_lag_analysis}
shows results of a more detailed analysis.
Figure~\ref{stability_variants}
shows that the importance-sampling based methods also perform better across all hyperparameters and are typically more robust.

\subsection{Multi-Task Training}

IMPALA's high data throughput and data efficiency allow us to train not only on one task but on multiple tasks in parallel with only a minimal change to the training setup. Instead of running the same task on all actors, we allocate a fixed number of actors to each task in the multi-task suite. Note, the model does not know which task it is being trained or evaluated on.

\subsubsection{DMLab-30}

\begin{table}[t]
\small
  \begin{center}
  \begin{tabular}{l@{\hspace{.22cm}}c@{\hspace{.22cm}}}
    \toprule
    \textbf{Model} &  \textbf{Test score}  \\ \midrule
      A3C, deep & 23.8\% \\
      IMPALA, shallow & 37.1\% \\
      IMPALA-Experts, deep & 44.5\% \\
      IMPALA, deep & 46.5\% \\
      IMPALA, deep, PBT & \textbf{49.4\%} \\
      IMPALA, deep, PBT, 8 learners & 49.1\% \\
      \bottomrule
  \end{tabular}
  \end{center}
\caption{Mean capped human normalised scores on DMLab-30. All models were evaluated on the test tasks with 500 episodes per task. The table shows the best score for each architecture.}
\label{dmlab30_table}
\end{table}

To test IMPALA's performance in a multi-task setting we use DMLab-30, a set of 30 diverse tasks built on DeepMind Lab. Among the many task types in the suite are visually complex environments with natural-looking terrain, instruction-based tasks with grounded language~\cite{hermann2017grounded}, navigation tasks, cognitive~\cite{leibo2018psychlab} and first-person tagging tasks featuring scripted bots as opponents. A detailed description of DMLab-30 and the tasks are available at \href{https://github.com/deepmind/lab}{github.com/deepmind/lab} and
\href{https://deepmind.com/dm-lab-30}{deepmind.com/dm-lab-30}.

We compare multiple variants of IMPALA with a distributed A3C implementation. Except for agents using population-based training (PBT) \cite{jaderberg2017pbt}, all agents are trained with hyperparameter sweeps across the same range given in Appendix~\ref{sec:hyperparameters}
. We report mean capped human normalised score where the score for each task is capped at 100\% (see Appendix~\ref{sec:ref_scores}
). Using mean capped human normalised score emphasises the need to solve multiple tasks instead of focusing on becoming super human on a single task. For PBT we use the mean capped human normalised score as fitness function and tune entropy cost, learning rate and RMSProp $\epsilon$. See Appendix~\ref{sec:pbt}
for the specifics of the PBT setup.

In particular, we compare the following agent variants. 
\textit{A3C, deep}, a distributed implementation with 210 workers (7 per task) featuring the deep residual network architecture (Figure~\ref{model_dmlab30} (Right)). \textit{IMPALA, shallow} with 210 actors and \textit{IMPALA, deep} with 150 actors both with a single learner. \textit{IMPALA, deep, PBT}, the same as \textit{IMPALA, deep}, but additionally using the PBT \cite{jaderberg2017pbt} for hyperparameter optimisation. Finally \textit{IMPALA, deep, PBT, 8 learners}, which utilises 8 learner GPUs to maximise learning speed. We also train IMPALA agents in an expert setting, \textit{IMPALA-Experts, deep}, where a separate agent is trained per task. In this case we did \textit{not} optimise hyperparameters for each task separately but instead across all tasks on which the 30 expert agents were trained.

\begin{figure}[t]
    \centering
    \caption{Performance of best agent in each sweep/population during training on the DMLab-30 task-set wrt. data consumed across all environments. IMPALA with multi-task training is not only faster, it also converges at higher accuracy with better data efficiency across all 30 tasks. The x-axis is data consumed by one agent out of a hyperparameter sweep/PBT population of 24 agents, total data consumed across the whole population/sweep can be obtained by multiplying with the population/sweep size.}
    \label{dmlab30_model_vs_a3c_experts}
    \includegraphics[width=0.85\columnwidth, trim={0 0 0 3.5cm},clip]{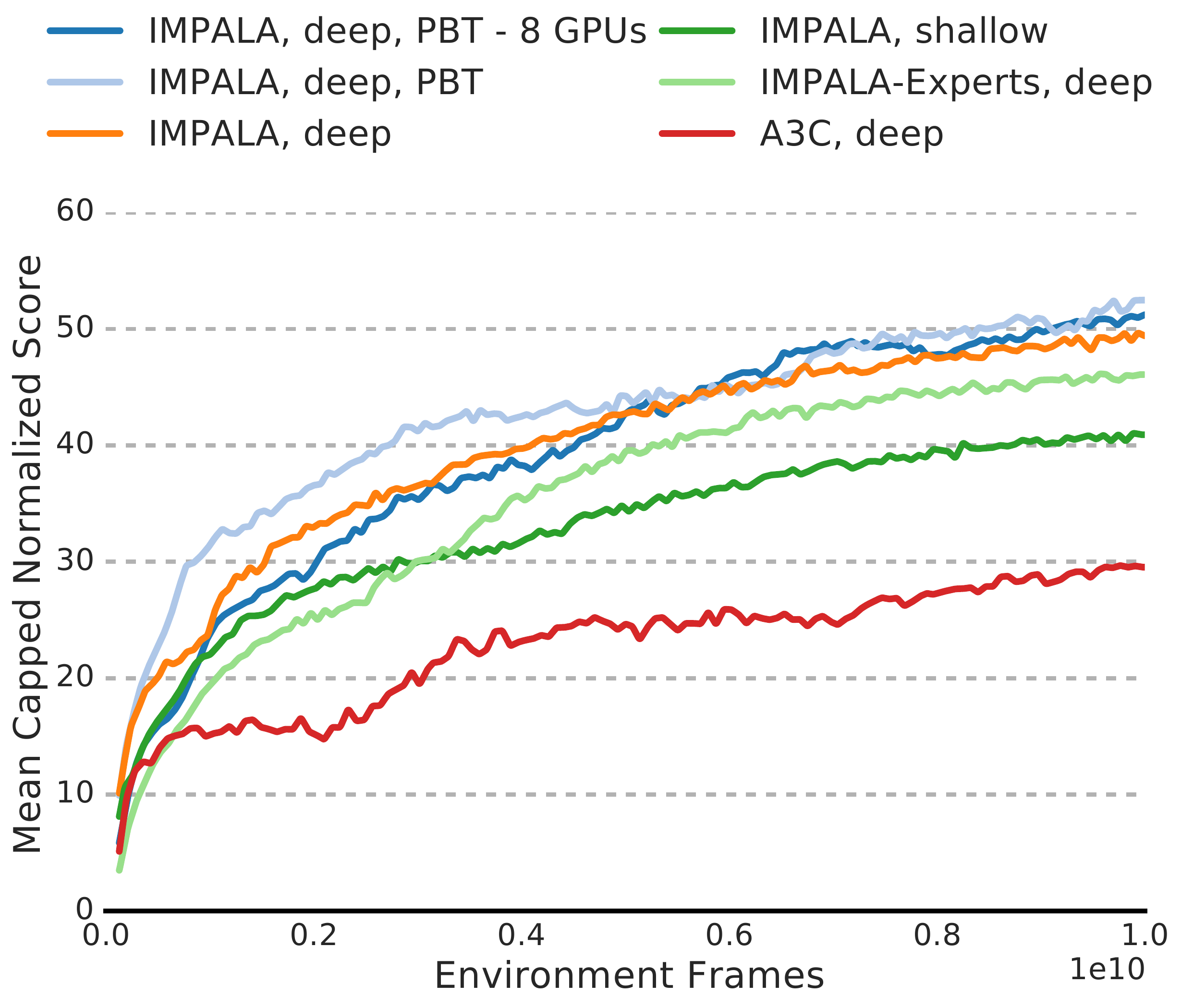}

    \includegraphics[width=0.85\columnwidth]{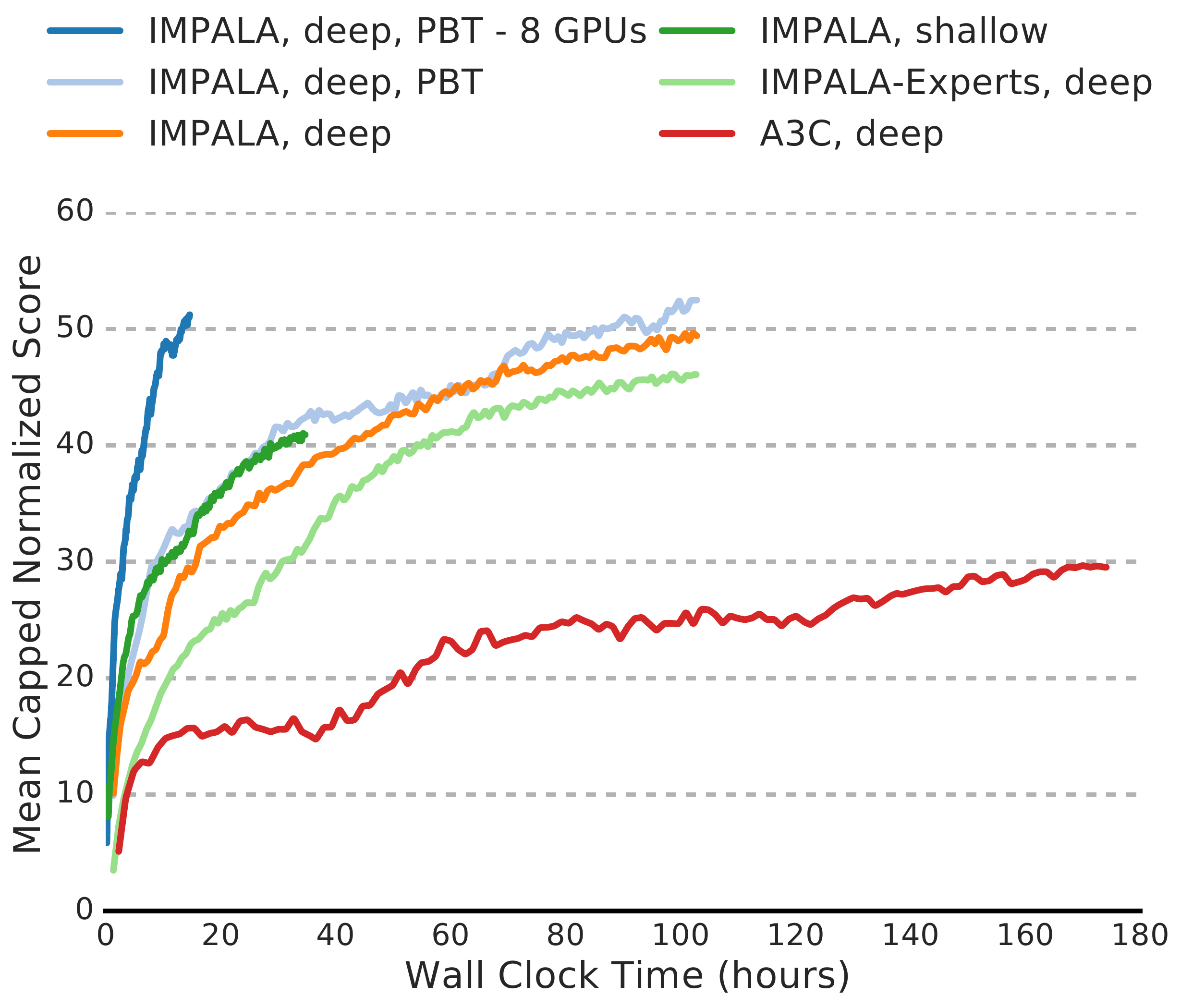}
    \caption{Performance on DMLab-30 wrt. wall-clock time. All models used the deep architecture (Figure~\ref{model_dmlab30}). The high throughput of IMPALA results in orders of magnitude faster learning.}
    \label{dmlab30_model_vs_a3c_experts_wallclock}
\end{figure}

Table~\ref{dmlab30_table} and Figure~\ref{dmlab30_model_vs_a3c_experts} show all variants of IMPALA performing much better than the deep distributed A3C. Moreover, the deep variant of IMPALA performs better than the shallow network version not only in terms of final performance but throughout the entire training. Note in Table~\ref{dmlab30_table} that \textit{IMPALA, deep, PBT, 8 learners}, although providing much higher throughput, reaches the same final performance as the 1 GPU \textit{IMPALA, deep, PBT} in the same number of steps.
Of particular importance is the gap between the \textit{IMPALA-Experts} which were trained on each task individually and \textit{IMPALA, deep, PBT} which was trained on all tasks at once.
As Figure~\ref{dmlab30_model_vs_a3c_experts} shows, the multi-task version is outperforms \textit{IMPALA-Experts} throughout training and the breakdown into individual scores in Appendix~\ref{sec:ref_scores}
shows positive transfer on tasks such as language tasks and laser tag tasks.

Comparing A3C to IMPALA with respect to wall clock time (Figure \ref{dmlab30_model_vs_a3c_experts_wallclock}) further highlights the scalability gap between the two approaches. IMPALA with 1 learner takes only around 10 hours to reach the same performance that A3C approaches after 7.5 days. Using 8 learner GPUs instead of 1 further speeds up training of the deep model by a factor of 7 to 210K frames/sec, up from 30K frames/sec.

\subsubsection{Atari}

The Atari Learning Environment (ALE) \cite{bellemare2013arcade} has been the testing ground of most recent deep reinforcement agents. Its 57 tasks pose challenging reinforcement learning problems including exploration, planning, reactive play and complex visual input. Most games feature very different visuals and game mechanics which makes this domain particularly challenging for multi-task learning.

We train IMPALA and A3C agents on each game individually and compare their performance using the deep network (without the LSTM) introduced in Section~\ref{sec:experiments}. We also provide results using a shallow network that is equivalent to the feed forward network used in \cite{A3C2016} which features three convolutional layers. The network is provided with a short term history by stacking the 4 most recent observations at each step. For details on pre-processing and hyperparameter setup please refer to Appendix~\ref{appendix:atari_experiments}
.

In addition to individual per-game experts, trained for 200 million frames with a fixed set of hyperparameters, we train an IMPALA Atari-57 agent---one agent, one set of weights---on all 57 Atari games at once for 200 million frames per game or a total of 11.4 billion frames. For the Atari-57 agent, we use population based training with a population size of 24 to adapt entropy regularisation, learning rate, RMSProp $\epsilon$ and the global gradient norm clipping threshold throughout training.

\begin{table}[ht]
\small
  \begin{center}
  \begin{tabular}{l@{\hspace{.22cm}}r@{\hspace{.22cm}}r@{\hspace{.22cm}}}
    \toprule
        \textbf{Human Normalised Return}         &  \textbf{Median}   & \textbf{Mean} \\
        \midrule
        A3C, shallow, experts           & 54.9\%             & 285.9\%    \\
        A3C, deep, experts              & 117.9\%            & 503.6\%    \\
        \midrule
        Reactor, experts                     & 187\%            & N/A    \\
        \midrule
        IMPALA, shallow, experts        & 93.2\%             & 466.4\%    \\
        IMPALA, deep, experts           & 191.8\%            & 957.6\%    \\
        \midrule
        IMPALA, deep, multi-task        & 59.7\%             & 176.9\%    \\
      \bottomrule
  \end{tabular}
  \end{center}
\caption{Human normalised scores on Atari-57. Up to 30 no-ops at the beginning of each episode. For a level-by-level comparison to ACKTR~\cite{acktr} and Reactor see~Appendix~\ref{tab:atari_individual_games}
.}
\label{atari57_table}
\end{table}

We compare all algorithms in terms of median human normalised score across all 57 Atari games. Evaluation follows a standard protocol, each game-score is the mean over 200 evaluation episodes, each episode was started with a random number of no-op actions (uniformly chosen from [1, 30]) to combat the determinism of the ALE environment.

As table~\ref{atari57_table} shows, IMPALA experts provide both better final performance and data efficiency than their A3C counterparts in the deep and the shallow configuration. As in our DeepMind Lab experiments, the deep residual network leads to higher scores than the shallow network, irrespective of the reinforcement learning algorithm used. Note that the shallow IMPALA experiment completes training over 200 million frames in less than one hour.

We want to particularly emphasise that \textit{IMPALA, deep, multi-task}, a single agent trained on all 57 ALE games at once, reaches 59.7\% median human normalised score. Despite the high diversity in visual appearance and game mechanics within the ALE suite, IMPALA multi-task still manages to stay competitive to \textit{A3C, shallow, experts}, commonly used as a baseline in related work. ALE is typically considered a hard multi-task environment, often accompanied by negative transfer between tasks \cite{rusu2016progressive}. To our knowledge, IMPALA is the first agent to be trained in a multi-task setting on all 57 games of ALE that is competitive with a standard expert baseline.

\section{Conclusion}
We have introduced a new highly scalable distributed agent, IMPALA, and a new off-policy learning algorithm, V-trace. 
With its simple but scalable distributed architecture, IMPALA can make efficient use of available compute at small and large scale.
This directly translates to very quick turnaround for investigating new ideas and opens up unexplored opportunities.

V-trace is a general off-policy learning algorithm that is more stable and robust compared to other off-policy correction methods for actor critic agents. We have demonstrated that IMPALA achieves better performance compared to A3C variants in terms of data efficiency, stability and final performance. We have further evaluated IMPALA on the new DMLab-30 set and the Atari-57 set. To the best of our knowledge, IMPALA is the first Deep-RL agent that has been successfully tested in such large-scale multi-task settings and it has shown superior performance compared to A3C based agents (49.4\% vs. 23.8\% human normalised score on DMLab-30). Most importantly, our experiments on DMLab-30 show that, in the multi-task setting, positive transfer between individual tasks lead IMPALA to achieve better performance compared to the expert training setting. We believe that IMPALA provides a simple yet scalable and robust framework for building better Deep-RL agents and has the potential to enable research on new challenges.

\clearpage
\section*{Acknowledgements}

We would like to thank Denis Teplyashin, Ricardo Barreira, Manuel Sanchez for their work improving the performance on DMLab-30 environments and Matteo Hessel, Jony Hudson, Igor Babuschkin, Max Jaderberg, Ivo Danihelka, Jacob Menick and David Silver for their comments and insightful discussions.

\bibliography{ms}
\bibliographystyle{icml2018}

\begin{bibunit}[icml2018]
\begin{appendices}
\counterwithin{figure}{section}
\counterwithin{table}{section}
\onecolumn
\icmltitle{Supplementary Material}

\section{Analysis of V-trace}\label{sec:analysis}
\subsection{V-trace operator}
Define the V-trace operator $\R$:
\beqa\label{eq:V-trace.operator}
\R V(x) &\eqdef& V(x) + \E_{\mu} \Big[\sum_{t\geq 0} \gamma^t \big(c_0\dots c_{t-1}\big) \rho_t\big(r_t+\gamma V(x_{t+1})-V(x_t)\big) \big|x_0=x, \mu \Big],
\eeqa
where the expectation $\E_{\mu}$ is with respect to the policy $\mu$ which has generated the trajectory $(x_t)_{t\geq 0}$, i.e., $x_0=x$, $x_{t+1}\sim p(\cdot|x_t, a_t)$, $a_t\sim \mu(\cdot|x_t)$.
Here we consider the infinite-horizon operator but very similar results hold for the $n$-step truncated operator.

\begin{theorem}\label{thm:V-trace.contraction}
Let $\rho_t = \min\big(\bar \rho, \frac{\pi(a_t|x_t)}{\mu(a_t|x_t)}\big)$ and $c_t = \min\big(\bar c, \frac{\pi(a_t|x_t)}{\mu(a_t|x_t)}\big)$ be truncated importance sampling weights, with $\bar \rho\geq \bar c$. Assume that there exists $\beta\in (0,1]$ such that $\E_{\mu}\rho_0\geq \beta$. Then the operator $\R$ defined by \eqref{eq:V-trace.operator} has a unique fixed point $V^{\pi_{\bar \rho}}$, which is the value function of the policy $\pi_{\bar \rho}$ defined by
 \beq\label{eq:pi_rho.apx}
 \pi_{\bar \rho}(a|x)\eqdef \frac{\min \big(\bar \rho \mu(a|x),\pi(a|x)\big)}{\sum_{b\in A}\min \big(\bar \rho \mu(b|x),\pi(b|x)\big)},
 \eeq
Furthermore, $\R$ is a $\eta$-contraction mapping in sup-norm, with 
$$\eta \eqdef \gamma^{-1} - (\gamma^{-1}-1) \E_{\mu} \big[ \sum_{t\geq 0} \gamma^{t} \big( \prod_{i=0}^{t-2} c_i\big)\rho_{t-1} \big] \leq 1-(1-\gamma)\beta<1.$$
\end{theorem}

\begin{remark} The truncation levels $\bar c$ and $\bar \rho$ play different roles in this operator:
\begin{itemize}
 \item $\bar \rho$ impacts the fixed-point of the operator, thus the policy $\pi_{\bar \rho}$ which is evaluated. For $\bar \rho=\infty$ (untruncated $\rho_t$) we get the value function of the target policy $V^{\pi}$, whereas for finite $\bar \rho$, we evaluate a policy which is in between $\mu$ and $\pi$ (and when $\rho$ is close to $0$, then we evaluate $V^{\mu}$). So the larger $\bar \rho$ the smaller the bias in off-policy learning. The variance naturally grows with $\bar \rho$. However notice that we do not take the product of those $\rho_t$ coefficients (in contrast to the $c_s$ coefficients) so the variance does not explode with the time horizon.
 \item $\bar c$ impacts the contraction modulus $\eta$ of $\R$ (thus the speed at which an online-algorithm like V-trace will converge to its fixed point $V^{\pi_{\bar \rho}}$). In terms of variance reduction, here is it really important to truncate the importance sampling ratios in $c_t$ because we take the product of those. Fortunately, our result says that for any level of truncation $\bar c$, the fixed point (the value function $V^{\pi_{\bar \rho}}$ we converge to) is the same: it does not depend on $\bar c$ but on $\bar \rho$ only.
\end{itemize}
\end{remark}

\begin{proof}
First notice that we can rewrite $\R$ as
\begin{align*}
\R V(x) &= (1- \E_\mu \rho_0) V(x) + \E_\mu \left [\sum_{t \ge 0} \gamma^t \Big( \prod_{s=0}^{t-1} c_s \Big ) \Big ( \rho_t r_t + \gamma [ \rho_t - c_{t} \rho_{t+1} ] V(x_{t+1}) \Big) \right ].
\end{align*}
Thus 
\begin{align*}
\R V_1(x) - \R V_2(x) &=  (1- \E_\mu \rho_0) \big[ V_1(x)-V_2(x)\big] + \E_\mu \left [\sum_{t \ge 0} \gamma^{t+1} \Big( \prod_{s=0}^{t-1} c_s \Big )  [ \rho_t - c_{t} \rho_{t+1} ] \big[ V_1(x_{t+1}) - V_2(x_{t+1})\big] \right ].\\
&= \E_\mu \left [\sum_{t \ge 0} \gamma^{t} \Big( \prod_{s=0}^{t-2} c_s \Big ) [ \underbrace{\rho_{t-1} - c_{t-1} \rho_{t}}_{\alpha_t} ] \big[ V_1(x_{t}) - V_2(x_{t}) \big] \right ],
\end{align*}
with the notation that $c_{-1}=\rho_{-1}=1$ and $\prod_{s=0}^{t-2} c_s = 1$ for $t=0$ and $1$.
Now the coefficients $(\alpha_t)_{t\geq 0 }$ are non-negative in expectation. Indeed, since $\bar \rho\geq \bar c$, we have
$$\E_\mu \alpha_t = \E \big[ \rho_{t-1} - c_{t-1} \rho_{t}\big] \geq \E_\mu \big[ c_{t-1}( 1- \rho_{t})\big] \geq 0,$$
since $\E_\mu \rho_{t} \leq \E_\mu \big[ \frac{\pi(a_t|x_t)}{\mu(a_t|x_t)}\big]  = 1$.
Thus $V_1(x) - V_2(x)$ is a linear combination of the values $V_1-V_2$ at other states, weighted by non-negative coefficients whose sum is
\beqan
& &\sum_{t \ge 0} \gamma^{t} \E_\mu \left [\Big( \prod_{s=0}^{t-2} c_s \Big ) [ \rho_{t-1} - c_{t-1} \rho_{t} ] \right ]\\
&=&  \sum_{t \ge 0} \gamma^{t} \E_\mu \left [\Big( \prod_{s=0}^{t-2} c_s \Big ) \rho_{t-1} \right ] - 
\sum_{t \ge 0} \gamma^{t} \E_\mu \left [\Big( \prod_{s=0}^{t-1} c_s \Big ) \rho_{t} \right ] \\
&=&  \sum_{t \ge 0} \gamma^{t} \E_\mu \left [\Big( \prod_{s=0}^{t-2} c_s \Big ) \rho_{t-1} \right ] - 
\gamma^{-1} \left(  \sum_{t \ge 0} \gamma^{t} \E_\mu \left [\Big( \prod_{s=0}^{t-2} c_s \Big ) \rho_{t-1} \right ] -1 \right) \\
&=& \gamma^{-1} - (\gamma^{-1} - 1)\underbrace{ \sum_{t \ge 0} \gamma^{t} \E_\mu \left [\Big( \prod_{s=0}^{t-2} c_s \Big ) \rho_{t-1} \right ]}_{\geq 1+\gamma\E_\mu\rho_0} \\
&\leq& 1-(1-\gamma) \E_\mu\rho_0 \\
&\leq& 1-(1-\gamma)\beta\\
&<&1.
\eeqan 

We deduce that $\| \R V_1(x)- \R V_2(x)  \| \leq \eta \|V_1 - V_2 \|_{\infty}$, with $\eta = \gamma^{-1} - (\gamma^{-1}-1) \sum_{t \ge 0} \gamma^{t} \E_\mu \left [\Big( \prod_{s=0}^{t-2} c_s \Big ) \rho_{t-1} \right ] \leq 1-(1-\gamma)\beta<1$, so $\R$ is a contraction mapping. Thus $\R$ possesses a unique fixed point. Let us now prove that this fixed point is $V^{\pi_{\bar \rho}}$. We have:
\beqan
& &\E_\mu \big[ \rho_t\big(r_t+\gamma V^{\pi_{\bar \rho}}(x_{t+1})-V^{\pi_{\bar \rho}}(x_t)\big)\big| x_t\Big] \\
&=& \sum_a \mu(a|x_t) \min\big(\bar \rho, \frac{\pi(a|x_t)}{\mu(a|x_t)} \big) \Big[r(x_t, a) + \gamma \sum_y p(y|x_t, a) V^{\pi_{\bar \rho}}(y) - V^{\pi_{\bar \rho}}(x_t)\Big] \\
&=& \underbrace{\sum_a \pi_{\bar \rho}(a|x_t) \Big[r(x_t, a) + \gamma \sum_y p(y|x_t, a) V^{\pi_{\bar \rho}}(y) - V^{\pi_{\bar \rho}}(x_t)\Big]}_{=0} \sum_b \min\big(\bar \rho \mu(b|x_t), \pi(b|x_t) \big) \\
&=& 0,
\eeqan
since this is the Bellman equation for $V^{\pi_{\bar \rho}}$. We deduce that $\R V^{\pi_{\bar \rho}} = V^{\pi_{\bar \rho}}$, thus $V^{\pi_{\bar \rho}}$ is the unique fixed point of $\R$.
\end{proof}

\subsection{Online learning}
\begin{theorem}\label{thm:online}
Assume a tabular representation, i.e.~the state and action spaces are finite. Consider a set of trajectories, with the $k^{th}$ trajectory $x_0, a_0, r_0, x_1, a_1, r_1, \dots$
generated by following $\mu$: $a_t\sim \mu(\cdot|x_t)$. For each state $x_s$ along this trajectory, update
\begin{align}
 V_{k+1}(x_s) & =  V_k(x_s) + \alpha_k(x_s) \sum_{t \ge s} \gamma^{t - s} \big (c_s \dots c_{t-1} \big) \rho_t  \big(r_t + \gamma V_k(x_{t+1}) - V_k(x_t)\big), 
 \end{align}
with $c_i=\min\big(\bar c,\frac{\pi(a_i|x_i)}{\mu(a_i|x_i)}\big)$, $\rho_i=\min\big(\bar \rho,\frac{\pi(a_i|x_i)}{\mu(a_i|x_i)}\big)$, $\bar \rho\geq \bar c$. Assume that (1) all states are visited infinitely often, and (2) the
stepsizes obey the usual Robbins-Munro conditions: for each state $x$, $\sum_k \alpha_k(x)=\infty$, $\sum_k \alpha_k^2(x) <\infty$. Then $V_k \to V^{\pi_{\bar \rho}}$ almost surely.
\end{theorem}

The proof is a straightforward application of the convergence result for stochastic approximation algorithms to the fixed point of a contraction operator, see e.g.~\cite{Dayan94,bertsekas1996neurodynamic,kushner2003}.

\subsection{On the choice of $q_s$ in policy gradient}
The policy gradient update rule (4) %\eqref{eq:off-policy.PG}
makes use of the coefficient $q_s = r_s + \gamma v_{s+1}$ as an estimate of  $Q^{\pi_{\bar \rho}}(x_s, a_s)$ built from the V-trace estimate $v_{s+1}$ at the next state $x_{s+1}$. The reason why we use $q_s$ instead of $v_s$ as target for our Q-value $Q^{\pi_{\bar \rho}}(x_s,a_s)$ is to make sure our estimate of the Q-value is as unbiased as possible, and the first requirement is that it is entirely unbiased in the case of perfect representation of the V-values. Indeed, assuming our value function is correctly estimated at all states, i.e.~$V=V^{\pi_{\bar \rho}}$, then we have $\E [q_s|x_s,a_s] = Q^{\pi_{\bar \rho}}(x_{s},a_s)$ (whereas we do not have this property for $v_t$). Indeed,
\begin{align*}
\E [q_s|x_s,a_s] &= r_s + \gamma \E\big[ V^{\pi_{\bar \rho}}(x_{s+1}) + \delta_{s+1} V^{\pi_{\bar \rho}} 
 + \gamma c_{s+1} \delta_{s+2} V^{\pi_{\bar \rho}} + \dots \big]\\
&= r_s + \gamma \E \big[ V^{\pi_{\bar \rho}}(x_{s+1}) \big] \\
&= Q^{\pi_{\bar \rho}}(x_{s},a_s)
\end{align*}
whereas 
\begin{align*}
\E [v_s|x_s,a_s] &= V^{\pi_{\bar \rho}}(x_s) + \rho_s\big( r_s+\gamma \E\big[V^{\pi_{\bar \rho}}(x_{s+1})\big] 
- V^{\pi_{\bar \rho}}(x_s)\big) + \gamma c_{s} \delta_{s+1} V^{\pi_{\bar \rho}} + \dots \\
&= V^{\pi_{\bar \rho}}(x_s) + \rho_s\big( r_s+\gamma \E\big[V^{\pi_{\bar \rho}}(x_{s+1})\big] - V^{\pi_{\bar \rho}}(x_s)\big)\\
&= V^{\pi_{\bar \rho}}(x_s)(1-\rho_s) + \rho_s Q^{\pi_{\bar \rho}}(x_s,a_s),
\end{align*}
which is different from $Q^{\pi_{\bar \rho}}(x_s,a_s)$ when $V^{\pi_{\bar \rho}}(x_s)\neq Q^{\pi_{\bar \rho}}(x_s,a_s)$.

\vfill

\section{Reference Scores}
\label{sec:ref_scores}

\begin{tabular}{l@{\hspace{.22cm}}r@{\hspace{.22cm}}r@{\hspace{.22cm}}r@{\hspace{.22cm}}r@{\hspace{.22cm}}}
\toprule
\textbf{Task $t$} &  \textbf{Human $h$} & \textbf{Random $r$} & \textbf{Experts} & \textbf{IMPALA}  \\ \midrule
\texttt{rooms\_collect\_good\_objects\_test} & 10.0 & 0.1 &  9.0 & 5.8 \\
\texttt{rooms\_exploit\_deferred\_effects\_test} & 85.7 & 8.5 & 15.6 & 11.0 \\
\texttt{rooms\_select\_nonmatching\_object} & 65.9 & 0.3 &  7.3 & 26.1 \\
\texttt{rooms\_watermaze} & 54.0 & 4.1 &  26.9 & 31.1 \\
\texttt{rooms\_keys\_doors\_puzzle} & 53.8 & 4.1 & 28.0 & 24.3 \\
\texttt{language\_select\_described\_object} & 389.5 & -0.1 & 324.6 & 593.1 \\
\texttt{language\_select\_located\_object} & 280.7 & 1.9 & 189.0 & 301.7 \\
\texttt{language\_execute\_random\_task} & 254.1 & -5.9 & -49.9 & 66.8 \\
\texttt{language\_answer\_quantitative\_question} & 184.5 & -0.3 & 219.4 & 264.0 \\
\texttt{lasertag\_one\_opponent\_large} & 12.7 & -0.2 & -0.2 & 0.3 \\
\texttt{lasertag\_three\_oponents\_large} & 18.6 & -0.2 & -0.1 & 4.1 \\
\texttt{lasertag\_one\_opponent\_small} & 18.6 & -0.1 & -0.1 & 2.5 \\
\texttt{lasertag\_three\_opponents\_small} & 31.5 & -0.1 & 19.1 & 11.3 \\
\texttt{natlab\_fixed\_large\_map} & 36.9 & 2.2 &  34.7 & 12.2 \\
\texttt{natlab\_varying\_map\_regrowth} & 24.4 & 3.0 &  20.7 & 15.9 \\
\texttt{natlab\_varying\_map\_randomized} & 42.4 & 7.3 &  36.1 & 29.0 \\
\texttt{skymaze\_irreversible\_path\_hard} & 100.0 & 0.1 & 13.6 & 30.0 \\
\texttt{skymaze\_irreversible\_path\_varied} & 100.0 & 14.4 & 45.1 & 53.6 \\
\texttt{pyschlab\_arbitrary\_visuomotor\_mapping} & 58.8 & 0.2 &  16.4 & 14.3 \\
\texttt{pyschlab\_continuous\_recognition} & 58.3 & 0.2 &  29.9 & 29.9 \\
\texttt{pyschlab\_sequential\_comparison} & 39.5 & 0.1 &  0.0 & 0.0 \\
\texttt{pyschlab\_visual\_search} & 78.5 & 0.1 &  0.0 & 0.0 \\
\texttt{explore\_object\_locations\_small} & 74.5 & 3.6 &  57.8 & 62.6 \\
\texttt{explore\_object\_locations\_large} & 65.7 & 4.7 & 37.0 & 51.1 \\
\texttt{explore\_obstructed\_goals\_small} & 206.0 & 6.8 &  135.2 & 188.8 \\
\texttt{explore\_obstructed\_goals\_large} & 119.5 & 2.6 & 39.5 & 71.0 \\
\texttt{explore\_goal\_locations\_small} & 267.5 & 7.7 &  209.4 & 252.5 \\
\texttt{explore\_goal\_locations\_large} & 194.5 & 3.1 & 83.1 & 125.3 \\
\texttt{explore\_object\_rewards\_few} & 77.7 & 2.1 &  39.8 & 43.2 \\
\texttt{explore\_object\_rewards\_many} & 106.7 & 2.4 & 58.7 & 62.6 \\
      \midrule
      Mean Capped Normalised Score: $\left(\sum_t \min\left[1, (s_t - r_t) / (h_t - r_t)\right]\right) / N$ & 100\% & 0\% & 44.5\% & 49.4\% \\
      \bottomrule
\end{tabular}
\captionof{table}{DMLab-30 test scores.}
\label{tab:dmlab_reference_scores}

\subsection{Final training scores on DMLab-30}
\label{sec:per_level_scores_dmlab30}

\begin{center}
\includegraphics[height=0.85\textheight]{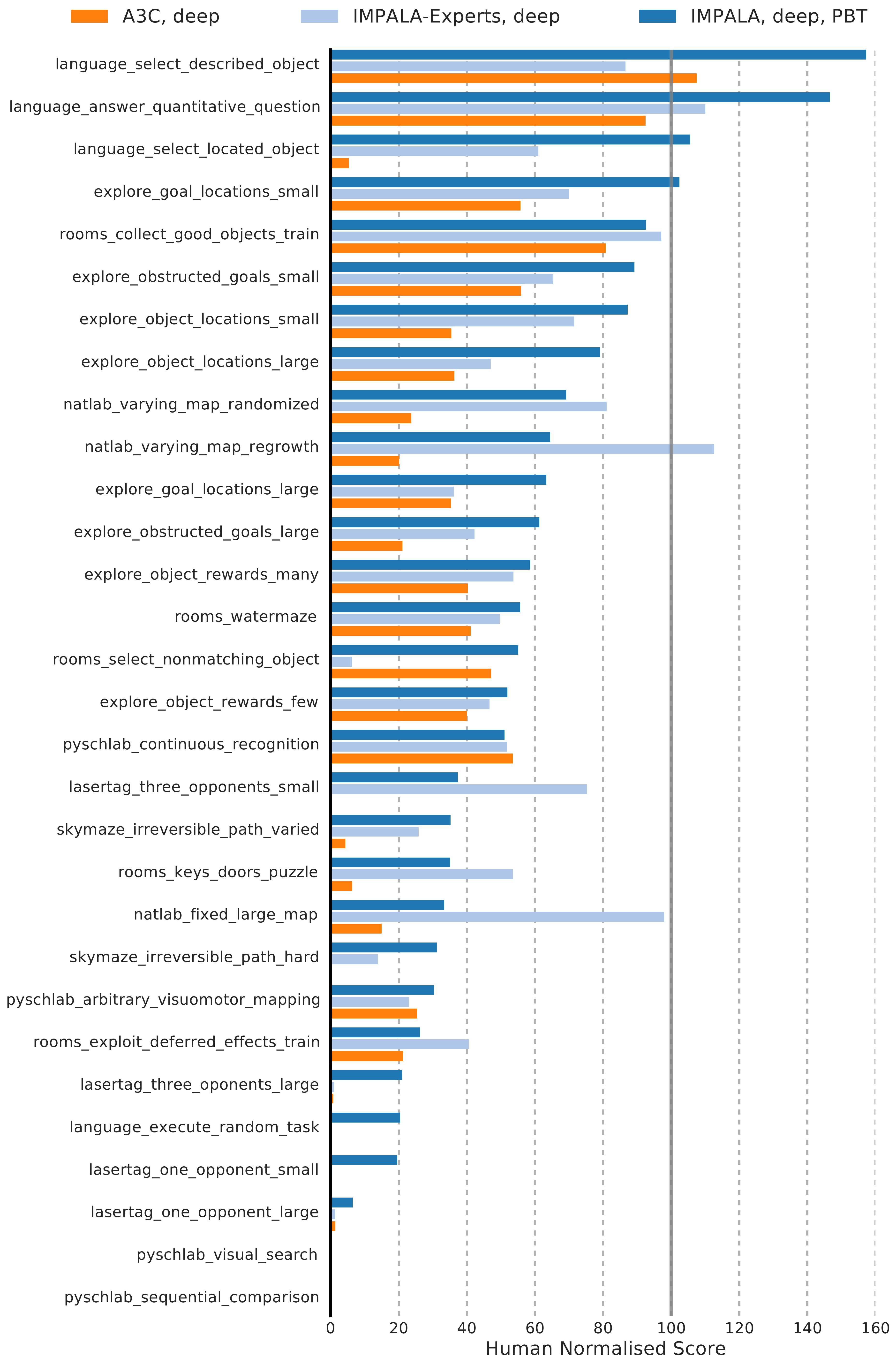}
\captionof{figure}{Human normalised scores across all DMLab-30 tasks.}
\end{center}
\vfill

\section{Atari Scores}
\label{sec:atari_scores}

\begin{center}
\small
\begin{tabular}{l@{\hspace{.26cm}}r@{\hspace{.26cm}}r@{\hspace{.26cm}}r@{\hspace{.26cm}}r@{\hspace{.26cm}}r@{\hspace{.26cm}}}
\toprule
                           &         ACKTR       &         The Reactor &         IMPALA (deep, multi-task) &         IMPALA (shallow) &         IMPALA (deep) \\
\midrule
        alien              &         3197.10     &         6482.10     &         2344.60                   &         1536.05          & \textbf{15962.10}     \\
        amidar             &         1059.40     &         833         &         136.82                    &         497.62           & \textbf{1554.79}      \\
        assault            &         10777.70    &         11013.50    &         2116.32                   &         12086.86         & \textbf{19148.47}     \\
        asterix            &         31583.00    &         36238.50    &         2609.00                   &         29692.50         & \textbf{300732.00}    \\
        asteroids          &         34171.60    &         2780.40     &         2011.05                   &         3508.10          & \textbf{108590.05}    \\
        atlantis           & \textbf{3433182.00} &         308258      &         460430.50                 &         773355.50        &         849967.50     \\
        bank\_heist        & \textbf{1289.70}    &         988.70      &         55.15                     &         1200.35          &         1223.15       \\
        battle\_zone       &         8910.00     & \textbf{61220}      &         7705.00                   &         13015.00         &         20885.00      \\
        beam\_rider        &         13581.40    &         8566.50     &         698.36                    &         8219.92          & \textbf{32463.47}     \\
        berzerk            &         927.20      &         1641.40     &         647.80                    &         888.30           & \textbf{1852.70}      \\
        bowling            &         24.30       & \textbf{75.40}      &         31.06                     &         35.73            &         59.92         \\
        boxing             &         1.45        &         99.40       &         96.63                     &         96.30            & \textbf{99.96}        \\
        breakout           &         735.70      &         518.40      &         35.67                     &         640.43           & \textbf{787.34}       \\
        centipede          &         7125.28     &         3402.80     &         4916.84                   &         5528.13          & \textbf{11049.75}     \\
        chopper\_command   &         N/A         & \textbf{37568}      &         5036.00                   &         5012.00          &         28255.00      \\
        crazy\_climber     &         150444.00   & \textbf{194347}     &         115384.00                 &         136211.50        &         136950.00     \\
        defender           &         N/A         &         113128      &         16667.50                  &         58718.25         & \textbf{185203.00}    \\
        demon\_attack      & \textbf{274176.70}  &         100189      &         10095.20                  &         107264.73        &         132826.98     \\
        double\_dunk       &         -0.54       & \textbf{11.40}      &         -1.92                     &         -0.35            &         -0.33         \\
        enduro             &         0.00        & \textbf{2230.10}    &         971.28                    &         0.00             &         0.00          \\
        fishing\_derby     &         33.73       &         23.20       &         35.27                     &         32.08            & \textbf{44.85}        \\
        freeway            &         0.00        & \textbf{31.40}      &         21.41                     &         0.00             &         0.00          \\
        frostbite          &         N/A         & \textbf{8042.10}    &         2744.15                   &         269.65           &         317.75        \\
        gopher             &         47730.80    & \textbf{69135.10}   &         913.50                    &         1002.40          &         66782.30      \\
        gravitar           &         N/A         & \textbf{1073.80}    &         282.50                    &         211.50           &         359.50        \\
        hero               &         N/A         & \textbf{35542.20}   &         18818.90                  &         33853.15         &         33730.55      \\
        ice\_hockey        &         -4.20       &         3.40        &         -13.55                    &         -5.25            & \textbf{3.48}         \\
        jamesbond          &         490.00      & \textbf{7869.20}    &         284.00                    &         440.00           &         601.50        \\
        kangaroo           &         3150.00     & \textbf{10484.50}   &         8240.50                   &         47.00            &         1632.00       \\
        krull              &         9686.90     &         9930.80     & \textbf{10807.80}                 &         9247.60          &         8147.40       \\
        kung\_fu\_master   &         34954.00    & \textbf{59799.50}   &         41905.00                  &         42259.00         &         43375.50      \\
        montezuma\_revenge &         N/A         & \textbf{2643.50}    &         0.00                      &         0.00             &         0.00          \\
        ms\_pacman         &         N/A         &         2724.30     &         3415.05                   &         6501.71          & \textbf{7342.32}      \\
        name\_this\_game   &         N/A         &         9907.20     &         5719.30                   &         6049.55          & \textbf{21537.20}     \\
        phoenix            &         133433.70   &         40092.20    &         7486.50                   &         33068.15         & \textbf{210996.45}    \\
        pitfall            & \textbf{-1.10}      &         -3.50       &         -1.22                     &         -11.14           &         -1.66         \\
        pong               &         20.90       &         20.70       &         8.58                      &         20.40            & \textbf{20.98}        \\
        private\_eye       &         N/A         & \textbf{15177.10}   &         0.00                      &         92.42            &         98.50         \\
        qbert              &         23151.50    &         22956.50    &         10717.38                  &         18901.25         & \textbf{351200.12}    \\
        riverraid          &         17762.80    &         16608.30    &         2850.15                   &         17401.90         & \textbf{29608.05}     \\
        road\_runner       &         53446.00    & \textbf{71168}      &         24435.50                  &         37505.00         &         57121.00      \\
        robotank           &         16.50       & \textbf{68.50}      &         9.94                      &         2.30             &         12.96         \\
        seaquest           &         1776.00     & \textbf{8425.80}    &         844.60                    &         1716.90          &         1753.20       \\
        skiing             &         N/A         &         -10753.40   & \textbf{-8988.00}                 &         -29975.00        &         -10180.38     \\
        solaris            &         2368.60     & \textbf{2760}       &         1160.40                   &         2368.40          &         2365.00       \\
        space\_invaders    &         19723.00    &         2448.60     &         199.65                    &         1726.28          & \textbf{43595.78}     \\
        star\_gunner       &         82920.00    &         70038       &         1855.50                   &         69139.00         & \textbf{200625.00}    \\
        surround           &         N/A         &         6.70        &         -8.51                     &         -8.13            & \textbf{7.56}         \\
        tennis             &         N/A         & \textbf{23.30}      &         -8.12                     &         -1.89            &         0.55          \\
        time\_pilot        &         22286.00    &         19401       &         3747.50                   &         6617.50          & \textbf{48481.50}     \\
        tutankham          & \textbf{314.30}     &         272.60      &         105.22                    &         267.82           &         292.11        \\
        up\_n\_down        & \textbf{436665.80}  &         64354.20    &         82155.30                  &         273058.10        &         332546.75     \\
        venture            &         N/A         & \textbf{1597.50}    &         1.00                      &         0.00             &         0.00          \\
        video\_pinball     &         100496.60   &         469366      &         20125.14                  &         228642.52        & \textbf{572898.27}    \\
        wizard\_of\_wor    &         702.00      & \textbf{13170.50}   &         2106.00                   &         4203.00          &         9157.50       \\
        yars\_revenge      & \textbf{125169.00}  &         102760      &         14739.41                  &         80530.13         &         84231.14      \\
        zaxxon             &         17448.00    &         25215.50    &         6497.00                   &         1148.50          & \textbf{32935.50}     \\
\bottomrule
\end{tabular}
\end{center}
\captionof{table}{Atari scores after 200M steps environment steps of training. Up to 30 no-ops at the beginning of each episode.}
\label{tab:atari_individual_games}

\section{Parameters}
\label{sec:variable_hyperparamters}

In this section, the specific parameter settings that are used throughout our experiments are given in detail.

\begin{center}
\begin{tabular}{l@{\hspace{.22cm}}l@{\hspace{.22cm}}l@{\hspace{.22cm}}}
    \toprule
    \textbf{Hyperparameter} & \textbf{Range} & \textbf{Distribution}  \\ \midrule
    Entropy regularisation & \texttt{[5e-5, 1e-2]} & Log uniform \\
    Learning rate & \texttt{[5e-6, 5e-3]} & Log uniform \\
    RMSProp epsilon $(\eps)$ regularisation parameter  & \texttt{[1e-1, 1e-3, 1e-5, 1e-7]} & Categorical \\
    \bottomrule
\end{tabular}
\captionof{table}{The ranges used in sampling hyperparameters across all experiments that used a sweep and for the initial hyperparameters for PBT. Sweep size and population size are 24. Note, the loss is \emph{summed} across the batch and time dimensions.}
\label{tab:hyperparameter_ranges}
\end{center}

\begin{center}
\begin{tabular}{l@{\hspace{.22cm}}r@{\hspace{.22cm}}}
\toprule
\textbf{Action} & \textbf{Native DeepMind Lab Action}  \\ \midrule
Forward & \texttt{[\ \ 0, 0,\ \ 0,\ \ 1, 0, 0, 0]} \\
Backward & \texttt{[\ \ 0, 0,\ \ 0, -1, 0, 0, 0]} \\
Strafe Left & \texttt{[\ \ 0, 0, -1,\ \ 0, 0, 0, 0]} \\
Strafe Right & \texttt{[\ \ 0, 0,\ \ 1,\ \ 0, 0, 0, 0]} \\
Look Left & \texttt{[-20, 0,\ \ 0,\ \ 0, 0, 0, 0]} \\
Look Right & \texttt{[ 20, 0,\ \ 0,\ \ 0, 0, 0, 0]} \\
Forward + Look Left & \texttt{[-20, 0,\ \ 0,\ \ 1, 0, 0, 0]} \\
Forward + Look Right & \texttt{[ 20, 0,\ \ 0,\ \ 1, 0, 0, 0]} \\
Fire & \texttt{[\ \ 0, 0,\ \ 0,\ \ 0, 1, 0, 0]} \\
      \bottomrule
\end{tabular}
\captionof{table}{Action set used in all tasks from the DeepMind Lab environment, including the DMLab-30 experiments.}
\label{tab:action_sets}
\end{center}

\subsection{Fixed Model Hyperparameters}
\label{sec:hyperparameters}
In this section, we list all the hyperparameters that were kept fixed across all experiments in the paper which are mostly concerned with observations specifications and optimisation. We first show below the reward pre-processing function that is used across all experiments using DeepMind Lab, followed by all fixed numerical values.

\begin{center}
\includegraphics[width=0.4\textwidth]{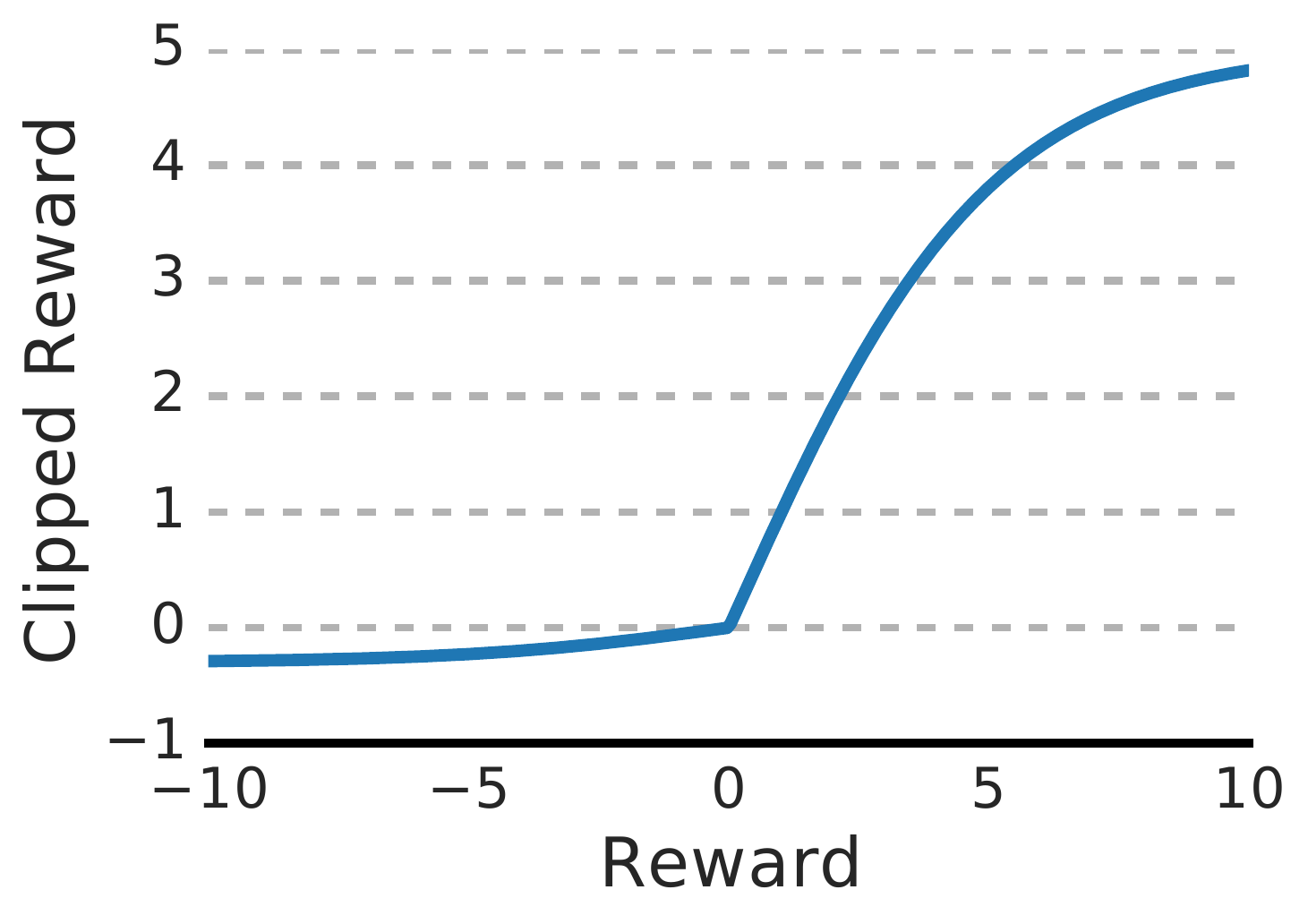}
\captionof{figure}{Optimistic Asymmetric Clipping - $0.3 \cdot \min(\tanh(reward), 0) + 5.0 \cdot \max(\tanh(reward), 0)$}
    \label{fig:asym_clipping}
\end{center}

\vspace{0.5cm}

\begin{center}
\begin{tabular}{l@{\hspace{.22cm}}r@{\hspace{.22cm}}}
\toprule
\textbf{Parameter} & \textbf{Value}  \\ \midrule
Image Width & 96 \\
Image Height & 72 \\
Action Repetitions & 4 \\
Unroll Length ($n$) & 100 \\
Reward Clipping & \\
\ \ \ \ - Single tasks & [-1, 1] \\
\ \ \ \ - DMLab-30, including experts & See Figure~\ref{fig:asym_clipping} \\
Discount ($\gamma$) & 0.99 \\
Baseline loss scaling & 0.5 \\
RMSProp momentum & 0.0 \\
Experience Replay (in Section~\ref{sec:vtrace_variants}
) & \\
\ \ \ \ - Capacity & 10,000 trajectories \\
\ \ \ \ - Sampling & Uniform \\
\ \ \ \ - Removal & First-in-first-out \\
      \bottomrule
\end{tabular}
\captionof{table}{Fixed model hyperparameters across all DeepMind Lab experiments.}
\label{tab:fixed_model_hyperparameters}
\end{center}
\vfill

\section{V-trace Analysis}
\label{sec:appendix_vtrace_analysis}

\subsection{Controlled Updates}
\label{sec:controlled_updates}
Here we show how different algorithms (On-Policy, No-correction, $\eps$-correction, V-trace) behave under varying levels of policy-lag between the actors and the learner.

\begin{center}
\includegraphics[width=\textwidth]{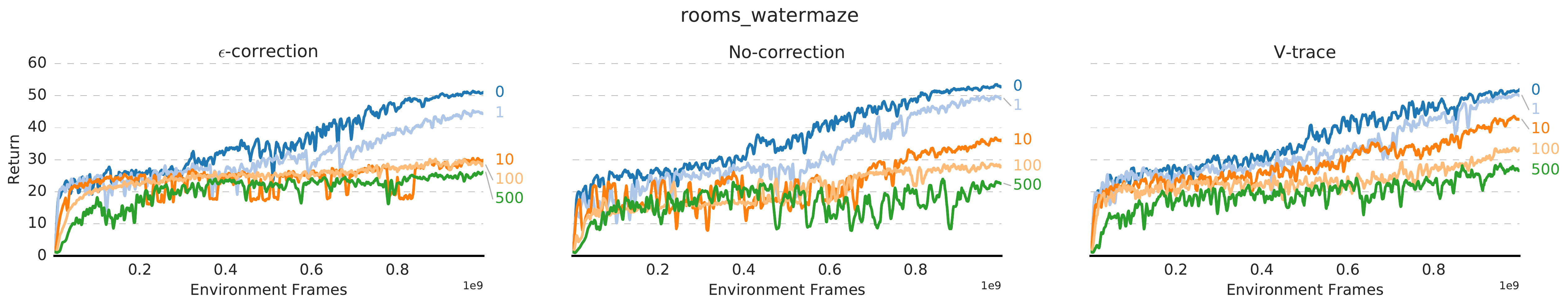}
\includegraphics[width=\textwidth]{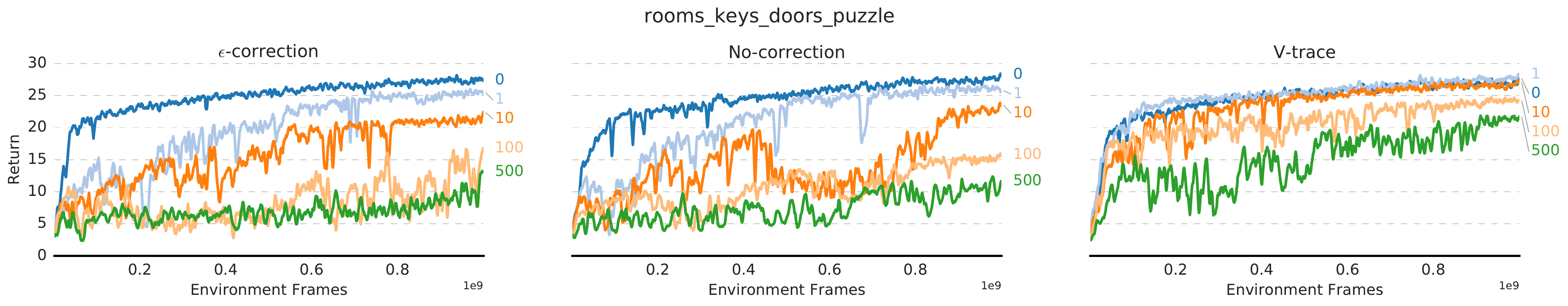}
\includegraphics[width=\textwidth]{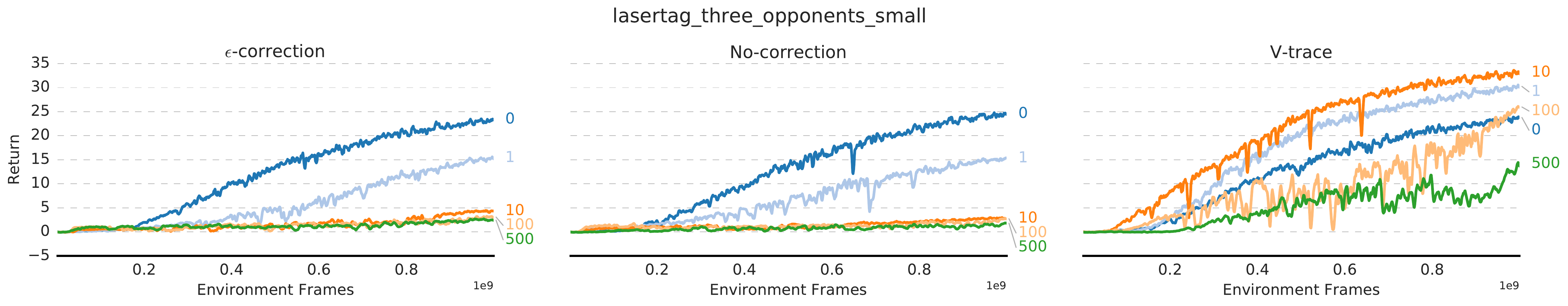}
\includegraphics[width=\textwidth]{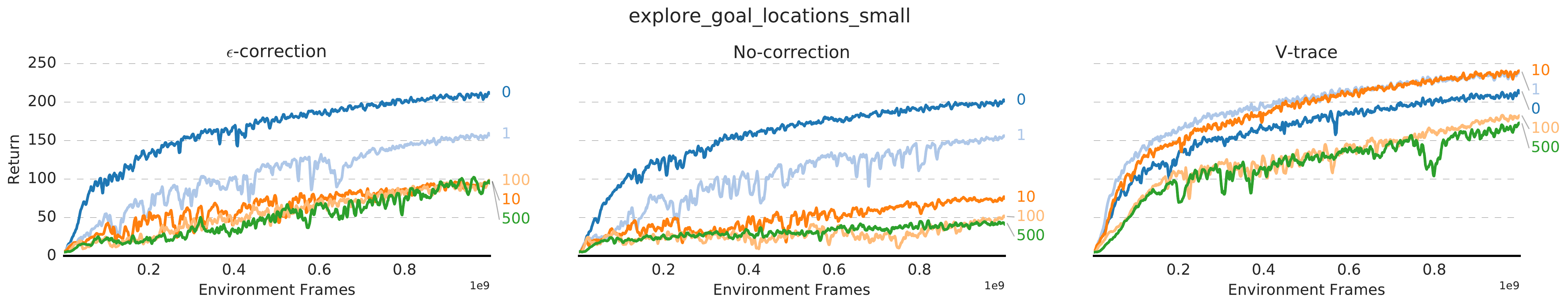}
\includegraphics[width=\textwidth]{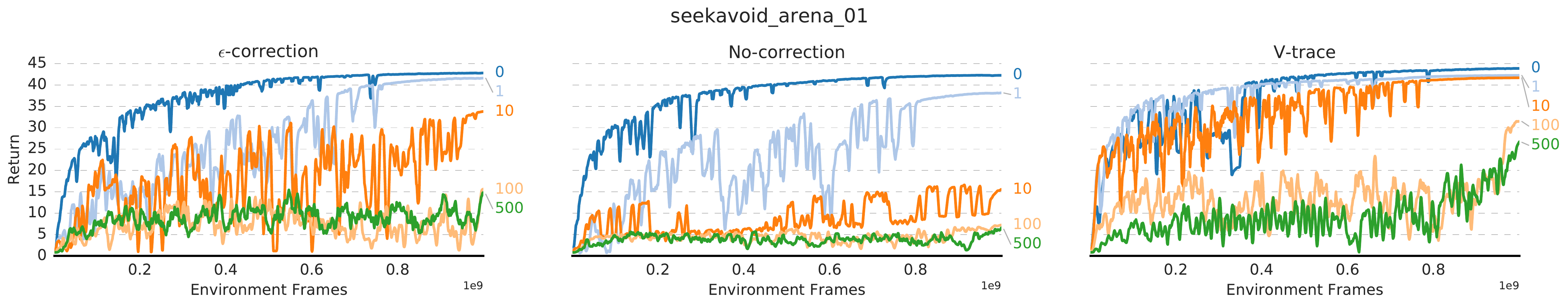}
\captionof{figure}{As the policy-lag (the number of update steps the actor policy is behind learner policy) increases, learning with V-trace is more robust compared to $\eps$-correction and pure on-policy learning.}
\label{fig:policy_lag_analysis}
\end{center}
\vfill

\subsection{V-trace Stability Analysis}

\begin{center}
    \includegraphics[width=.98\textwidth]{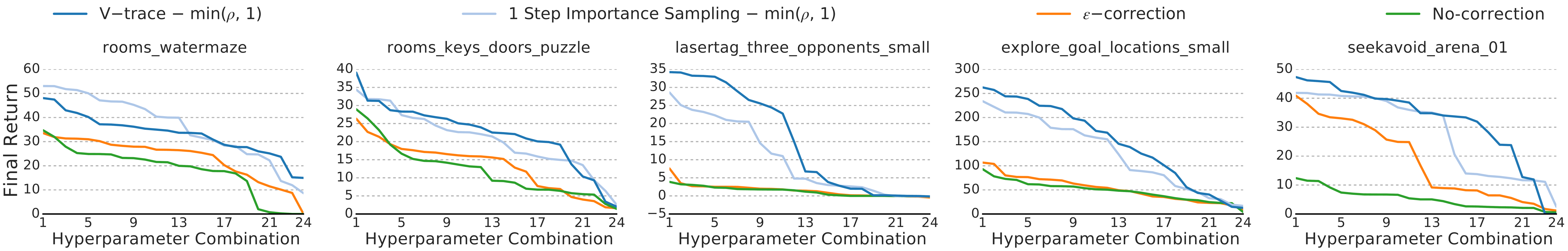}
    \captionof{figure}{Stability across hyper parameter combinations for different off-policy correction variants using replay. V-trace is much more stable across a wide range of parameter combinations compared to $\eps$-correction and pure on-policy learning.}
    \label{stability_variants}
\end{center}

\subsection{Estimating the State Action Value for Policy Gradient}
\label{sec:appendix_q_estimation}
We investigated different ways of estimating the state action value function used to estimate advantages for the policy gradient calculation.
The variant presented in the main section of the paper uses the V-trace corrected value function $v_{s+1}$ to estimate $q_s = r_s + \gamma v_{s+1}$.
Another possibility is to use the actor-critic baseline $V(x_{s+1})$ to estimate $q_s = r_s + \gamma V(x_{s+1})$.
Note that the latter variant does not use any information from the current policy rollout to estimate the policy gradient and relies on an accurate estimate of the value function.
We found the latter variant to perform worse both when comparing the top 3 runs and an average over all runs of the hyperparameter sweep as can be see in figures~\ref{q_estimation_top3} and \ref{q_estimation_all}.

\begin{center}
    \includegraphics[width=.98\textwidth]{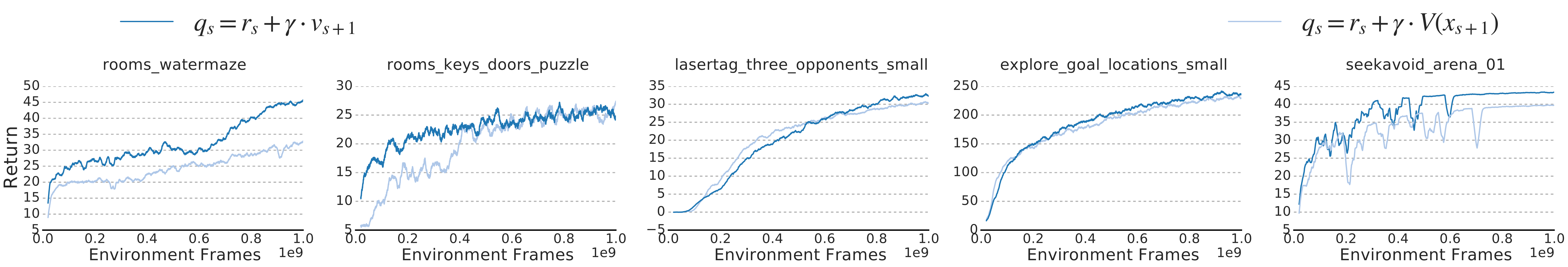}
    \captionof{figure}{Variants for estimation of state action value function - average over top 3 runs.}
    \label{q_estimation_top3}
\end{center}

\begin{center}
    \includegraphics[width=.98\textwidth]{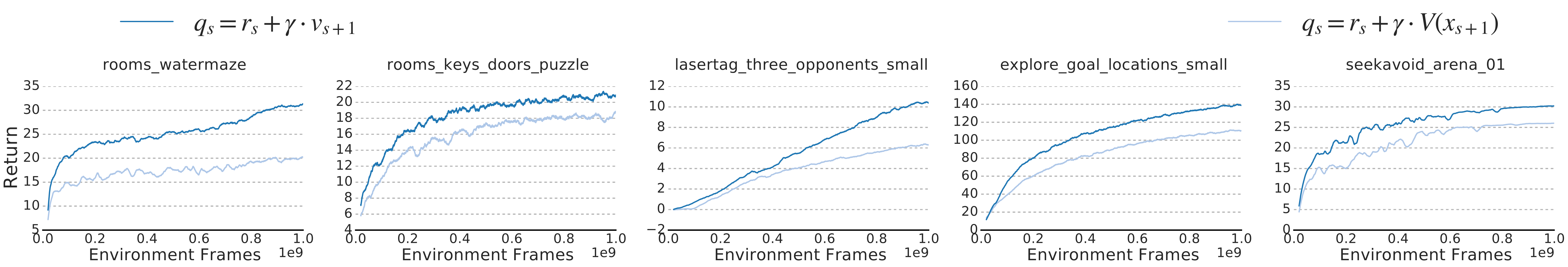}
    \captionof{figure}{Variants for estimation of state action value function - average over all runs.}
    \label{q_estimation_all}
\end{center}

\section{Population Based Training}
\label{sec:pbt}

For Population Based Training we used a ``burn-in" period of 20 million frames where no evolution is done. This is to stabilise the process and to avoid very rapid initial adaptation which hinders diversity. After collecting 5,000 episode rewards in total, the mean capped human normalised score is calculated and a random instance in the population is selected. If the score of the selected instance is more than an absolute 5\% higher, then the selected instance weights and parameters are copied.

No matter if a copy happened or not, each parameter (RMSProp epsilon, learning rate and entropy cost) is permuted with 33\% probability by multiplying with either $1.2$ or $1/1.2$. This is different from \cite{jaderberg2017pbt} in that our multiplication is unbiased where they use a multiplication of $1.2$ or $.8$. We found that diversity is increased when the parameters are permuted even if no copy happened.

We reconstruct the learning curves of the PBT runs in Figure~\ref{dmlab30_model_vs_a3c_experts}
by backtracking through the ancestry of copied checkpoints for selected instances.

\begin{figure}[H]
\begin{center}
    \includegraphics[scale=.3]{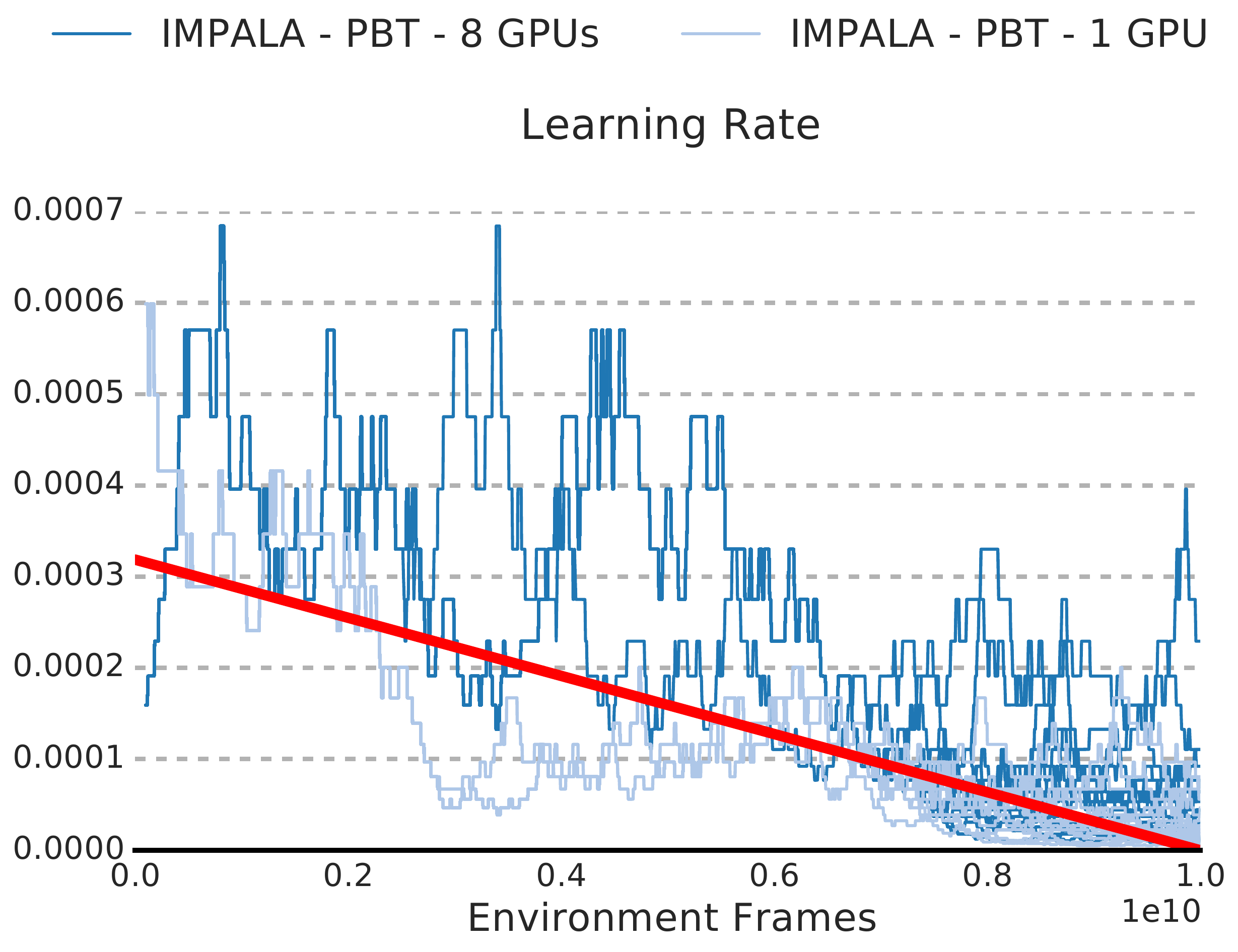}
\end{center}
\caption{Learning rate schedule that is discovered by the PBT \cite{jaderberg2017pbt} method compared against the linear annealing schedule of the best run from the parameter sweep (red line).}
\label{fig:pbt_learning_rate_schedule}
\end{figure}

\section{Atari Experiments}
\label{appendix:atari_experiments}

All agents trained on Atari are equipped only with a feed forward network and pre-process frames in the same way as described in \cite{A3C2016}.
When training experts agents, we use the same hyperparameters for each game for both IMPALA and A3C. These hyperparameters are the result of tuning A3C with a shallow network on the following games:
\texttt{breakout}, \texttt{pong}, \texttt{space\_invaders}, \texttt{seaquest}, \texttt{beam\_rider}, \texttt{qbert}. Following related work, experts use game-specific action sets.

The multi-task agent was equipped with a feed forward residual network (see Figure~\ref{model_dmlab30}
). The learning rate, entropy regularisation, RMSProp $\epsilon$ and gradient clipping threshold were adapted through population based training.
To be able to use the same policy layer on all Atari games in the multi-task setting we train the multi-task agent on the full Atari action set consisting of 18 actions.

Agents were trained using the following set of hyperparameters:

\begin{table}[H]
\begin{center}
\begin{tabular}{l@{\hspace{.22cm}}l@{\hspace{.22cm}}}
\toprule
\textbf{Parameter} & \textbf{Value}  \\
\midrule
Image Width & 84 \\
Image Height & 84 \\
Grayscaling & Yes \\
Action Repetitions & 4 \\
Max-pool over last N action repeat frames & 2 \\
Frame Stacking & 4 \\
End of episode when life lost & Yes \\
Reward Clipping & [-1, 1] \\
Unroll Length ($n$) & 20 \\
Batch size & 32 \\
Discount ($\gamma$) & 0.99 \\
Baseline loss scaling & 0.5 \\
Entropy Regularizer & 0.01 \\
RMSProp momentum & 0.0 \\
RMSProp $\epsilon$ & 0.01 \\
Learning rate & 0.0006 \\
Clip global gradient norm & 40.0 \\
Learning rate schedule & Anneal linearly to 0 \\
                       & From beginning to end of training. \\
Population based training (only multi-task agent) & \\
\ \ \ \ - Population size & 24 \\
\ \ \ \ - Start parameters & Same as DMLab-30 sweep \\
\ \ \ \ - Fitness & Mean capped human normalised scores \\
                  & $\left(\sum_l \min\left[1, (s_t - r_t) / (h_t - r_t)\right]\right) / N$ \\
\ \ \ \ - Adapted parameters & Gradient clipping threshold \\
                             & Entropy regularisation \\
                             & Learning rate \\
                             & RMSProp $\epsilon$ \\
      \bottomrule
\end{tabular}
\end{center}
\caption{Hyperparameters for Atari experiments.}
\label{tab:fixed_model_hyperparameters_atari}
\end{table}

\putbib[ms]
\end{appendices}
\end{bibunit}

\end{document}